\documentclass[10pt,twocolumn,letterpaper]{article}

\usepackage{cvpr}
\usepackage{times}
\usepackage{epsfig}
\usepackage{graphicx}
\usepackage{amsmath}
\usepackage{amssymb}
\usepackage{booktabs}
\usepackage{float}
\usepackage{longtable}
\usepackage[pagebackref=false,breaklinks=true,letterpaper=true,colorlinks,bookmarks=false]{hyperref}

\cvprfinalcopy
\def\cvprPaperID{****}

\ifcvprfinal\fi

\begin{document}

\title{Predicting Deep Neural Network Training Outcomes\\from Early Training Telemetry}

\author{Ranjita Naik\\
Georgia Institute of Technology\\
{\tt\small rnaik36@gatech.edu}
\and
Anh D. Nguyen\\
Georgia Institute of Technology\\
{\tt\small anguyen650@gatech.edu}
\and
Pankaj Kumar Singh\\
Georgia Institute of Technology\\
{\tt\small psingh448@gatech.edu}
}

\maketitle

\begin{abstract}
Large hyperparameter sweeps for deep neural networks routinely spend most of
their compute on configurations that are effectively doomed from the first
few epochs. We study whether a single training run's own early telemetry --
per-epoch loss, training accuracy, gradient signal-to-noise ratio,
weight-norm growth, and a one-time activation-saturation snapshot --
together with its sampled hyperparameters, is sufficient to predict that
run's eventual outcome. Critically, we make this prediction without
reference to any other concurrently or previously observed run: many
existing early-stopping methods
work by comparing a run against its peers in the same sweep, whereas we ask
what can be inferred from a run in isolation. We consider three prediction
targets -- final test accuracy (regression), whether a run will land in the
upper half of its cohort (relative classification), and whether a run will
terminate in a training-dynamics failure, including divergence to NaN
(failure prediction) -- and ask specifically whether gradient- and
weight-level internal telemetry adds information beyond the loss and
accuracy curves that most practitioners already monitor. We generate
23{,}788 unique training runs spanning six architecture/dataset combinations
(ResNet-18, a compact convolutional network, and a two-layer multilayer
perceptron, each trained on CIFAR-10 and Fashion-MNIST) using a two-phase
hyperparameter sampling procedure whose second phase deliberately
concentrates draws near the estimated success/failure boundary. On a
permanently held-out 20\% partition of hyperparameter configurations, never
touched during model development, gradient-boosted trees using only the
first five epochs of telemetry and the four sampled hyperparameters achieve
$R^2=0.92$--$0.99$ for final-accuracy regression and ROC-AUC$=0.983$--$0.998$
for relative classification across all six domains, with useful prediction
already available after a single epoch. A paired, hyperparameter-free
ablation isolating gradient and weight-norm signals from ordinary
loss/accuracy curves shows that internal telemetry provides a statistically
consistent improvement over curves alone in every domain and task, though
the practical size of that improvement varies considerably by domain.
Predictors transfer with little degradation between architecturally similar
domains, but transfer that also crosses dataset boundaries is limited mainly
by a mismatch in absolute accuracy scale rather than a loss of the
underlying rank relationship: Spearman rank correlation and domain-relative
classification remain substantially more stable under transfer than raw
regression error. We discuss these results as evidence for early-training
telemetry as a source of decision-support signal for compute allocation,
rather than as a basis for fully automatic termination of training runs, and
report the conditions -- boundary-focused sampling, short 15-epoch runs, two
image datasets, three architectures -- under which the findings should be
interpreted.
\end{abstract}

\section{Introduction}
\label{sec:intro}

Hyperparameter sweeps for deep neural networks are wasteful in a specific,
predictable way: many sampled configurations are either obviously going to
converge well or obviously going to fail, often visibly so within the
first few epochs. Multi-fidelity methods such as successive
halving~\cite{jamieson2016} and Hyperband~\cite{li2017} exploit this by
running many configurations briefly and discarding the worst-performing
fraction, and production tuning services use similar comparison-based
stopping rules~\cite{golovin2017}. These methods answer a different
question than the one we study here: they decide whether to continue a run
\emph{relative to the other runs in the sweep}, rather than asking what a
run's own early trajectory already implies about its outcome. A rule based
on a run's own telemetry needs no concurrent cohort to compare against,
and could flag a doomed single-run job or supply an independent signal
alongside whatever comparison-based scheduling a sweep already uses.

This paper studies that second question: how much can be inferred about a
training run's eventual outcome from its own telemetry, in isolation, over
its first few epochs. We focus on two properties of a run's internal
training dynamics that are cheap to record but not routinely used for this
purpose -- the signal-to-noise ratio of its parameter gradients and the
growth of its weight norm -- and test whether they add predictive value
beyond the loss and training-accuracy curves that are already universally
logged. If a run is going to diverge or plateau, we would expect this to
leave a trace in how consistently its gradients point in a given direction
and how quickly its weights are moving, potentially before that
instability is visible in the loss curve itself. We test this hypothesis
with a controlled, hyperparameter-free ablation rather than relying on
correlational evidence alone.

We study three prediction targets, formalized in
Section~\ref{sec:formulation}: final test accuracy (regression), whether a
run lands in the upper half of its cohort (relative classification, since
``good'' accuracy is domain-dependent), and whether a run terminates in a
training-dynamics failure, including outright divergence to NaN, not just
a low final score (failure prediction). We keep relative performance and
failure risk as separate tasks because collapsing them into one accuracy
cutoff would be arbitrary across domains with very different achievable
accuracies.

To study this where the success/failure boundary is well populated --
rather than trivially easy to predict because most runs are clearly fine
or clearly broken -- we generate 23{,}788 unique training runs across six
architecture/dataset combinations (ResNet-18~\cite{he2016}, a compact
convolutional network, and a two-layer multilayer perceptron, each trained
on CIFAR-10 and Fashion-MNIST), using a two-phase sampling procedure whose
second phase deliberately concentrates draws near the estimated boundary
(Section~\ref{sec:data}). We evaluate every headline result on a
permanently held-out 20\% partition of configurations never touched during
model development, in addition to a repeated-holdout protocol used during
development itself (Section~\ref{sec:protocol}).

\textbf{Findings.} Gradient-boosted trees using five epochs of telemetry and
the sampled hyperparameters predict final accuracy with $R^2=0.92$--$0.99$
and relative classification with ROC-AUC$=0.983$--$0.998$ across all six
domains on the frozen holdout, and useful prediction is already available
after a single epoch. A controlled, hyperparameter-free ablation shows that
gradient- and weight-level telemetry improves prediction beyond loss and
accuracy curves alone in every domain and task tested, though the size of
that improvement ranges from practically negligible to substantial
depending on the domain. Predictors transfer with little degradation
between architecturally similar domains; transfer that also crosses dataset
boundaries degrades in raw regression error mainly because of a mismatch in
absolute accuracy scale between datasets, while rank-based and
domain-relative metrics hold up much better. One combination -- transferring
a predictor trained on a convolutional architecture to a multilayer
perceptron on CIFAR-10 -- transfers noticeably worse than every other
architecture pair, in a way specific to that dataset rather than to the
multilayer perceptron architecture in general.

\textbf{Limits of these claims.} Our evaluation cohort is deliberately
enriched near the success/failure boundary, so reported error rates and
class balances describe this boundary-heavy population, not an arbitrary
practitioner's sweep (Section~\ref{sec:limitations}). All runs are capped at
15 epochs on two image datasets with three architecture families, and we do
not vary the optimizer or learning-rate schedule beyond what the sampled
hyperparameters already cover. Strong early prediction is evidence that a
run's outcome is often foreshadowed early, not evidence that it is safe to
terminate runs automatically without human oversight; we return to this
distinction in Section~\ref{sec:ethics}.

\textbf{Our contributions are:}
\begin{itemize}
\item A controlled, hyperparameter-free experimental design that isolates
the marginal contribution of gradient- and weight-level internal telemetry
from both ordinary loss/accuracy curves and from sampled hyperparameters,
evaluated across six architecture/dataset combinations.
\item A leakage-safe, multi-horizon evaluation protocol with a permanently
frozen holdout partition, used for every headline result reported.
\item An empirical study of failure prediction that treats outright
training divergence as a distinct, directly predicted outcome rather than
folding it into a single accuracy threshold.
\item A cross-architecture and cross-dataset transfer analysis that
separates the effect of scale mismatch (which degrades raw regression
error) from the effect of a genuinely different learned relationship
(which degrades rank-based and relative metrics), across all
one-factor-at-a-time domain pairs among our six domains.
\end{itemize}

\section{Related Work}
\label{sec:related}

\textbf{Learning-curve extrapolation.} A learning curve for a single
training run can be extrapolated to predict its eventual value without
comparing it to any other run. Domhan~\etal~\cite{domhan2015} fit a
weighted combination of parametric curve models to a run's own partial
validation loss/accuracy history and extrapolate forward, terminating runs
whose extrapolated performance is unlikely to beat the best run seen so
far. Klein~\etal~\cite{klein2017} replace the parametric curve model with a
Bayesian neural network that produces a full predictive distribution over
the extrapolated curve rather than a point estimate. Both approaches
operate purely on the scalar loss or accuracy trajectory; neither
incorporates gradient- or weight-level internal telemetry, and neither
predicts outright divergence as a distinct target rather than as a
degenerate case of low final accuracy.

\textbf{Multi-fidelity hyperparameter optimization.} A different family of
methods allocates a fixed compute budget across many configurations by
running each briefly and discarding the worst-performing fraction.
Successive halving~\cite{jamieson2016} and its bandit-based generalization
Hyperband~\cite{li2017} repeatedly halve a population of configurations
based on their current performance; BOHB~\cite{falkner2018} combines this
scheduling strategy with a Bayesian optimization model to make the
sampling of new configurations itself more sample-efficient. These methods
are comparison-based: a configuration survives or is discarded relative to
the other configurations currently being evaluated in the same sweep,
which requires a population of concurrently running jobs. Our setting is
complementary -- we ask what can be inferred from a single run's own
telemetry in isolation, which does not require a concurrent cohort and
could in principle supply an additional signal to a comparison-based
scheduler rather than replacing it.

\textbf{Early stopping and run termination.} Production black-box
optimization services implement simpler run-termination heuristics in
practice; the median stopping rule used in Google
Vizier~\cite{golovin2017}, for instance, terminates a run if its current
performance falls behind the median of previously completed runs at the
same point in training. Like the multi-fidelity methods above, this is a
comparison against other runs (here, historical rather than concurrent)
rather than a judgment based on a run's own trajectory alone.

\textbf{Performance prediction for neural architectures.}
Baker~\etal~\cite{baker2018} predict a network's final validation accuracy
from a combination of its early learning-curve values, its architecture,
and its hyperparameters, and use the prediction to accelerate architecture
search; this is the closest prior work to our setting, since it also
predicts a single run's outcome from its own early curves and
hyperparameters rather than by comparison to other runs. We extend this
line of work in two ways: we add gradient- and weight-level internal
optimization telemetry alongside the curves and hyperparameters, and we
show with a controlled, hyperparameter-free ablation
(Section~\ref{sec:ablation}) that this telemetry carries information the
curves alone do not; and we additionally predict outright divergence to NaN
as a distinct outcome, rather than only a continuous final-performance
score. Unterthiner~\etal~\cite{unterthiner2020} take a different approach,
predicting a trained network's test accuracy directly from its final
learned weights rather than from telemetry recorded during training, which
answers a related but distinct question (evaluating an already-trained
network rather than forecasting a network that is still training).
NAS-Bench-101~\cite{ying2019} is a standard benchmark for this class of
performance-prediction problem in the architecture-search setting, though
it evaluates architecture-level rather than hyperparameter-level variation
and does not include internal optimization telemetry.

\textbf{Zero-cost neural architecture search proxies.} A separate line of
work scores \emph{untrained} architectures using statistics computed from a
single mini-batch of gradients, in order to rank candidate architectures
without training any of them to convergence.
Abdelfattah~\etal~\cite{abdelfattah2021} survey and unify several such
zero-cost proxies. These methods use gradient statistics for a
fundamentally different purpose than ours: they compare untrained
architectures to each other at initialization, whereas we track how
gradient statistics evolve over several epochs of actual training for a
single, already-selected architecture and hyperparameter configuration, and
use them to forecast that specific run's outcome rather than to rank
architectures.

\textbf{Training-dynamics and optimization telemetry.} The idea that
gradient statistics reflect the state of an ongoing optimization is well
established in the large-batch training literature:
McCandlish~\etal~\cite{mccandlish2018} define a gradient noise scale from
the relationship between gradient variance and batch size and use it to
inform batch-size selection during training, which is conceptually related
to (though algorithmically distinct from) the gradient signal-to-noise
ratio we track here. To our knowledge, this class of internal optimization
signal has not previously been evaluated as a predictor of a run's
eventual success, failure, or final accuracy in a controlled comparison
against ordinary loss/accuracy curves, which is the gap this paper
addresses.

\textbf{How this work differs.} Unlike multi-fidelity and comparison-based
methods (successive halving, Hyperband, BOHB, median stopping), our study
never compares a run against its peers: every prediction is a function of
one run's own telemetry and hyperparameters. Unlike prior single-run
performance prediction~\cite{baker2018}, we add gradient- and weight-level
internal optimization telemetry to curves and hyperparameters, and test its
marginal contribution with a controlled, hyperparameter-free ablation
rather than relying on the curves and hyperparameters alone. Unlike
zero-cost NAS proxies, which score untrained architectures at
initialization, we track how gradient statistics evolve over several
epochs of actual training for one already-selected configuration, and use
them to forecast that specific run's outcome -- including outright
numerical divergence, treated as its own predicted outcome rather than a
degenerate case of low accuracy.

\section{Problem Formulation}
\label{sec:formulation}

We consider a training run $r$ produced by sampling a hyperparameter
configuration $h(r)$ and training a fixed architecture on a fixed dataset
for a fixed number of epochs. At the end of epoch $e$, the run has produced
a vector of per-epoch telemetry statistics $\tau_e(r)$ (defined precisely
in Section~\ref{sec:data}) covering training loss, training accuracy, a
gradient signal-to-noise statistic, and a weight-norm growth statistic,
together with a one-time activation-saturation snapshot recorded at a fixed
early epoch. For an observation horizon $k$, we define the available
information for run $r$ as
\[
\mathcal{I}_k(r) = \big(h(r),\ \tau_1(r), \tau_2(r), \dots, \tau_k(r)\big),
\]
that is, the sampled hyperparameters together with every telemetry
statistic recorded up to and including epoch $k$, and nothing recorded
after epoch $k$ or at the run's completion. All three prediction tasks
below are functions of $\mathcal{I}_k(r)$ alone; no task ever conditions on
telemetry from another run or on any information recorded after epoch $k$.

\textbf{Final-accuracy regression.} Predict $y_{\text{acc}}(r) \in [0,100]$,
the run's final test accuracy, for runs that reach a terminal state with a
defined accuracy (Section~\ref{sec:protocol} specifies the eligible cohort
precisely, since a run that diverges numerically has no such value).

\textbf{Relative (high-performance) classification.} Predict
$y_{\text{rel}}(r) = \mathbb{1}[y_{\text{acc}}(r) > m]$, where $m$ is the
median final accuracy computed on a training partition of the same domain.
Using a domain-relative threshold rather than one fixed absolute accuracy
value is necessary because the six domains in this study have very
different achievable accuracies (Section~\ref{sec:results}); a single
global threshold would make the task trivial in some domains and
impossible in others.

\textbf{Failure prediction.} Predict $y_{\text{fail}}(r) =
\mathbb{1}[\text{run } r \text{ terminates in a training-dynamics failure}]$,
where a training-dynamics failure is either (a) the run completing its full
training schedule with final accuracy below a fixed, domain-specific
usability threshold set independently of any prediction task, or (b) the
run's loss diverging numerically before completion. At horizon $k$, only
runs still active through epoch $k$ are eligible for this task, since a run
that already failed before $k$ could not have produced a horizon-$k$
prediction in a live deployment.

We deliberately keep relative classification and failure prediction as two
separate tasks rather than one combined threshold, because they answer two
different questions with two different natural thresholds: relative
classification asks whether a run is likely to be among the better runs in
its sweep, while failure prediction asks whether a run is at risk of not
producing a usable model at all, including the qualitatively different
failure mode of numerical divergence.

\section{Data Generation and Telemetry Collection}
\label{sec:data}

\subsection{Architectures, datasets, and run counts}

We train three architecture families -- ResNet-18~\cite{he2016}, a compact
custom convolutional network, and a two-hidden-layer multilayer perceptron
-- each from scratch on two image classification datasets, CIFAR-10 and
Fashion-MNIST, giving six architecture/dataset domains. Every run is capped
at 15 epochs, long enough for both successful and failing training
dynamics to become apparent while keeping the overall study affordable at
this scale, since our goal is predicting outcomes early rather than
obtaining the best possible final accuracy for any individual run. Table~\ref{tab:accounting}
gives the exact run count retained in each domain after data cleaning
(Section~\ref{sec:dataquality}); the study totals 23{,}788 unique training
runs.

\begin{table}[t]
\centering
\footnotesize
\setlength{\tabcolsep}{3.2pt}
\begin{tabular}{lccrr}
\toprule
Domain & Bar (\%) & Raw & Removed & Final \\
\midrule
C10 / ResNet-18   & 45 & 3{,}998 & 0   & 3{,}998 \\
C10 / SmallCNN    & 40 & 4{,}000 & 0   & 4{,}000 \\
C10 / MLP         & 30 & 4{,}000 & 0   & 4{,}000 \\
FMNIST / ResNet-18 & 80 & 3{,}990 & 200 & 3{,}790 \\
FMNIST / SmallCNN  & 78 & 4{,}000 & 0   & 4{,}000 \\
FMNIST / MLP       & 70 & 4{,}000 & 0   & 4{,}000 \\
\midrule
Total & -- & 23{,}988 & 200 & \textbf{23{,}788} \\
\bottomrule
\end{tabular}
\caption{Run counts and the usability threshold for each domain (C10 =
CIFAR-10, FMNIST = Fashion-MNIST). The threshold is the fixed final-accuracy
level below which a completed run counts as a training-dynamics failure,
chosen before generating any data (Section~\ref{sec:usability}). The 200
removed rows are a duplicate export, resolved as described in
Section~\ref{sec:dataquality}. C10/ResNet-18 and FMNIST/ResNet-18 fall 2 and
10 runs short of the 4{,}000-run target reached elsewhere; every raw row in
both domains carries a genuine terminal status, so the shortfall reflects
sampled configurations for which no output was ever recorded, not a failure
that was silently dropped (Appendix~\ref{app:dataaccounting}).}
\label{tab:accounting}
\end{table}

\subsection{Hyperparameter sampling}
\label{sec:sampling}

For each domain, we sample four hyperparameters per run: learning rate,
batch size, weight decay, and whether batch normalization is used. Sampling
proceeds in two phases. The first phase draws all four hyperparameters
broadly across a wide range (full ranges are given in
Appendix~\ref{app:setup}), producing a population of runs with a broad,
largely arbitrary spread of outcomes. We then fit a lightweight classifier
to the first phase's outcomes that estimates the probability of a
training-dynamics failure as a function of the four sampled
hyperparameters. The second phase uses this estimated failure probability
to concentrate roughly half of its draws near the estimated
success/failure boundary, while continuing to sample the other half
broadly to keep the domain well represented overall.

This two-phase design directly addresses a specific risk in any
hyperparameter-outcome prediction study: if nearly every sampled
configuration is either obviously successful or obviously broken, the
prediction task becomes close to trivial and uninformative about harder,
boundary-adjacent cases. Because the second phase deliberately oversamples
the boundary region, the class balance and error rates reported in
Section~\ref{sec:results} describe this boundary-enriched population,
rather than the distribution an arbitrary practitioner's sweep would
produce; we return to the interpretive consequences of this design choice
in Sections~\ref{sec:discussion} and~\ref{sec:limitations}.

\subsection{Usability thresholds and training-dynamics failure}
\label{sec:usability}

A completed run is labeled a training-dynamics failure if it finishes all
15 epochs with final accuracy below a fixed, domain-specific usability
threshold (Table~\ref{tab:accounting}), and a run whose loss diverges
numerically at any point is labeled a failure regardless of epoch. The six
thresholds were fixed before any data was generated and were never
estimated from evaluation data; each was chosen to sit clearly above chance
performance but well below a well-tuned run's achievable accuracy, roughly
half of a well-tuned run's accuracy on the harder CIFAR-10 datasets and
closer to a well-tuned run's accuracy on the easier Fashion-MNIST datasets
(where a literal half-of-ceiling threshold would rarely be crossed at all).
Full derivation and terminal-status breakdown are given in
Appendix~\ref{app:dataaccounting}.

\subsection{Telemetry signals}
\label{sec:telemetry}

Each run records, at every epoch $e$, its training loss, training
accuracy, a gradient signal-to-noise statistic, and a weight-norm growth
statistic, plus a one-time activation-saturation snapshot recorded at a
fixed early epoch (by default, epoch 2). We summarize the three internal
optimization signals here and give the full formal definitions in
Appendix~\ref{app:telemetry}.

\emph{Gradient signal-to-noise ratio} measures, for a fixed random sample
of parameter coordinates drawn once per run, how consistently the gradient
at each sampled coordinate points in the same direction across a bounded
number of mini-batches within an epoch, versus how much it fluctuates; a
high value indicates a consistent, low-noise gradient direction, and a low
value indicates a noisy one. \emph{Weight-norm growth} measures the
relative change in the $L_2$ norm of all trainable parameters, summed
across the network, relative to its value at initialization; large or
rapidly growing values can indicate an optimization trajectory that is
moving away from initialization unusually quickly, a pattern associated
with instability. \emph{Activation saturation} measures the fraction of
penultimate-layer activations that are at or below zero, averaged over a
small number of batches; a highly saturated network has fewer active units
contributing gradient signal. Because activation saturation is measured
once rather than tracked every epoch, it is excluded from the feature set
at the earliest observation horizon (before its one measurement epoch has
occurred) and included from that epoch onward, following the same
leakage-safe rule applied to every other feature (Section~\ref{sec:protocol}).

\subsection{Data-quality issues found and resolved}
\label{sec:dataquality}

Two data-identity issues were identified and corrected before any
predictive modeling was performed. First, the naive per-row identifier
assigned during data collection is reused across the two sampling phases
and is therefore not a safe key for deduplication; we instead deduplicate
on a hash of the sampled hyperparameter tuple, which is invariant to
sampling phase, and verified that no domain contains two runs that share a
hyperparameter hash but disagree on their random training seed (which
would indicate a genuine identity collision rather than a naming
artifact). Second, one domain (Fashion-MNIST/ResNet-18) had two overlapping
data exports covering the same 200 run identifiers, one at finer
granularity and one at coarser granularity: 40 of the corresponding rows
were byte-identical between the two exports, and the remaining 160 shared
identifiers but disagreed on their recorded telemetry values, consistent
with ordinary GPU/cuDNN non-determinism across separate executions of the
same configuration, and concentrated disproportionately near the
second-phase success/failure boundary. Rather than resolving the 160
disagreeing rows by an arbitrary tie-breaking rule, we removed the entire
redundant finer-granularity export, which is the 200-row removal shown in
Table~\ref{tab:accounting}.

\section{Prediction Tasks and Evaluation Protocol}
\label{sec:protocol}

\subsection{Leakage-safe, multi-horizon evaluation}

We evaluate every prediction task at four observation horizons, $k \in
\{1,2,3,5\}$ epochs. At horizon $k$, a run is only eligible for evaluation
if it produced real telemetry through epoch $k$, and every feature used at
that horizon is built strictly from information available at or before
epoch $k$ (Section~\ref{sec:formulation}); no feature at horizon $k$ ever
uses telemetry recorded after epoch $k$, or the run's final outcome. This
rule applies uniformly, including to the once-measured activation-saturation
signal, which is simply absent from the feature set at horizons before its
measurement epoch rather than treated as a special case.

Because \emph{final-accuracy regression} and \emph{relative classification}
require a well-defined final accuracy, a run whose loss diverges to NaN is
not eligible for either task; the eligible cohort for these two tasks is
therefore restricted to runs that reach a terminal state with a real
accuracy value. Such diverged runs are still eligible for \emph{failure
prediction}, which is defined over a broader cohort of runs with any known
terminal status. At a given horizon $k$, failure prediction further
restricts its eligible cohort to runs still observable through epoch $k$ --
a run that already diverged before epoch $k$ is excluded, since no
horizon-$k$ prediction could have been made for it in a real deployment.
The relative-classification median threshold, and the eligibility rule
that removes already-failed runs from the failure-prediction cohort at
horizon $k$, are the only two places any task's definition depends on
information from other runs; the trained model itself, at prediction time,
still only ever sees one run's own telemetry and hyperparameters.

\subsection{Feature sets and the internal-telemetry ablation}
\label{sec:featuresets}

Our richest feature representation, used for the headline results in
Section~\ref{sec:results}, engineers a set of summary statistics (first
value, last value, change from first to last, mean, standard deviation,
minimum, maximum, area under the curve, slope, and sign-reversal count) for
each of the four per-epoch telemetry signals at each observation horizon,
combined with the four sampled hyperparameters (log-transformed where
appropriate; see Appendix~\ref{app:setup}) and the once-measured activation
saturation value.

To isolate the specific contribution of internal optimization telemetry,
we additionally construct two feature sets that never include the sampled
hyperparameters: a \emph{curves-only} set containing only raw per-epoch
loss and training-accuracy values through the observation horizon, and a
\emph{curves-plus-internal} set that adds the raw per-epoch gradient
signal-to-noise and weight-norm growth values to the same curves. Both are
evaluated under the identical repeated train/test splits described in
Section~\ref{sec:evalsplits}, so the comparison between them is a paired
comparison on matched folds rather than two independently noisy numbers.
This pairing, not the full feature-set ablation ladder reported in
Section~\ref{sec:ablation}, is the controlled test of whether internal
telemetry adds information beyond curves: the full ladder also includes
feature sets that add the sampled hyperparameters, which by themselves
already carry substantial predictive information (Section~\ref{sec:ablation})
and would otherwise confound a claim about telemetry specifically.

\subsection{Models}

We compare simple baselines (predicting the training-fold majority class,
mean, or median), linear models (logistic regression for classification,
ridge regression for regression), random forests, and gradient-boosted
trees, which is our primary model throughout. To confirm that our results
are not an artifact of using tree ensembles specifically, we additionally
train a small feed-forward neural network directly on the telemetry
features, with dropout and early stopping, for one domain
(CIFAR-10/ResNet-18); extending this comparison to all six domains is left
for future work given the added computational cost of training and tuning
a neural predictor separately per domain (Section~\ref{sec:limitations}).
Every cross-domain comparison and every transfer result in
Section~\ref{sec:results} uses gradient-boosted trees.

\subsection{Repeated grouped holdout for model development}
\label{sec:evalsplits}

During model development -- comparing feature sets, models, and horizons --
we use repeated grouped holdout evaluation on an \emph{open} 80\% partition
of each domain's configurations (the remaining 20\% is permanently reserved
and never used for development; see Section~\ref{sec:frozenholdout}). For
each (endpoint, horizon, feature set, model) combination, we draw ten
train/test splits at a 25\% test fraction, grouping by hyperparameter
configuration so that, if any configuration were ever repeated, all of its
runs would stay in the same partition; no domain in this study in fact
contains repeated training seeds for identical configurations, so this
grouped split is equivalent in practice to an ordinary configuration-level
random split, though it remains a safeguard worth verifying rather than
assuming. We summarize the resulting ten per-split values with their mean
and a 95\% confidence interval using the Nadeau--Bengio correction, which
accounts for the fact that repeated splits drawn from overlapping data are
positively correlated and would otherwise produce an artificially narrow,
overconfident interval; we give the full correction formula in
Appendix~\ref{app:ci}. These intervals describe split-to-split variability
within our fixed set of sampled configurations; they are not a
generalization bound to some new, independently drawn population of
hyperparameter configurations.

\subsection{Frozen holdout for final evaluation}
\label{sec:frozenholdout}

The repeated-holdout protocol above is a variance estimate over one fixed
sweep, not a true test of generalization to configurations untouched by
development, since each of the ten splits still draws its test fraction
from the same open 80\% partition used throughout development. To obtain a
genuine one-shot generalization estimate, we additionally reserve, before
any model development begins, a fixed 20\% of each domain's hyperparameter
configurations as a permanently frozen holdout. The full pipeline proceeds
in order: generate the first sampling phase, fit the boundary-focused
sampler on its outcomes, generate the second sampling phase, deduplicate
and clean the combined data, freeze 20\% of configurations, develop
predictors, ablations, and horizon curves using only the remaining open
80\%, and finally evaluate once, at the end, on the frozen 20\%. The
boundary-focused sampler used during data generation is fit only on
first-phase outcomes and is not itself an outcome predictor, so it cannot
leak information into this final evaluation step. Every headline number in
Section~\ref{sec:results} reporting a single point estimate (as opposed to
a confidence interval) is this frozen-holdout, one-shot score. One
limitation of this design is discussed in
Section~\ref{sec:limitations}: because the second sampling phase
deliberately concentrates draws near the success/failure boundary in a
four-dimensional hyperparameter space populated by only a few thousand
samples per domain, some frozen-partition configurations likely sit close
to, though never identical to, an open-partition configuration, which is a
plausible partial explanation for how strong prediction already is at the
shortest observation horizon (Section~\ref{sec:horizonresults}).

\subsection{Transfer evaluation}
\label{sec:transferprotocol}

To test whether a predictor generalizes across architectures and datasets,
rather than only within the domain it was trained on, we evaluate every
ordered pair of the six domains. For a source domain and a target domain,
we fit the primary gradient-boosted model on the source domain's open 80\%
partition and evaluate it, with no retraining or fine-tuning, on the target
domain's frozen 20\% partition, so every transfer number is computed on
data the target-domain model never saw during training or development. We
report results separately for pairs that differ in exactly one of
\{architecture, dataset\} (Section~\ref{sec:crossarch}
and~\ref{sec:crossdataset}) and for pairs that differ in both simultaneously
(Appendix~\ref{app:transfer}), so that a low or high transfer score can be
attributed to a single factor wherever possible. Because CIFAR-10 and
Fashion-MNIST runs occupy very different absolute accuracy ranges
(per-domain median final accuracies range from 40.9 to 89.2 across our six
domains), we define relative classification for transfer evaluation using
each domain's own median rather than one shared absolute threshold, so that
transfer performance measures whether the learned relationship
generalizes rather than whether the two domains happen to share a common
accuracy scale. For regression transfer we report both raw error, which
directly reflects any such scale mismatch, and Spearman rank correlation,
which does not.

\section{Experimental Results}
\label{sec:results}

\subsection{In-domain prediction}
\label{sec:indomain}

At the five-epoch observation horizon, on the frozen holdout, gradient
boosting reaches $R^2 \geq 0.92$ for final-accuracy regression and
ROC-AUC $\geq 0.983$ for relative classification in every one of the six
domains (Table~\ref{tab:headline}); failure-prediction ROC-AUC is
$\geq 0.991$ in every domain over the same horizon. For CIFAR-10/ResNet-18
specifically, gradient boosting achieves MAE $=3.23$, $R^2=0.964$, Spearman
$=0.981$ for regression (778 eligible runs), against a mean-baseline MAE of
$26.9$ and $R^2 \approx 0$; ROC-AUC $=0.996$ and balanced accuracy $=0.960$
for relative classification (778 runs), against a majority-baseline
ROC-AUC of $0.500$; and ROC-AUC $=0.993$ for failure prediction (782 runs).
The regression and classification cohorts are slightly smaller than the
failure-prediction cohort because the former two require a well-defined
final accuracy that a diverged run never produces
(Section~\ref{sec:protocol}); the $R^2=0.92$--$0.99$ range reported in the
abstract accordingly describes runs that reached a terminal state with a
real accuracy value, not diverged runs. The feed-forward neural network
predictor trained on the same features for this domain achieves MAE
$=3.17$, $R^2=0.966$, and classification ROC-AUC $=0.997$ on the same
frozen holdout -- close enough to gradient boosting (MAE $3.17$ vs.\
$3.23$) that a single one-shot comparison cannot meaningfully distinguish
the two model families, which is consistent with our reading of this
result as evidence that the predictive signal is present in the telemetry
itself rather than an artifact of one particular model class.

\begin{table}[t]
\centering
\footnotesize
\begin{tabular}{lccc}
\toprule
Domain & Reg.\ $R^2$ & Clf.\ AUC & Fail.\ AUC \\
\midrule
C10 / ResNet-18   & 0.964 & 0.996 & 0.993 \\
C10 / SmallCNN    & 0.956 & 0.998 & 0.991 \\
C10 / MLP         & 0.990 & 0.998 & 0.999 \\
FMNIST / ResNet-18 & 0.923 & 0.983 & 0.991 \\
FMNIST / SmallCNN  & 0.948 & 0.992 & 0.993 \\
FMNIST / MLP       & 0.982 & 0.995 & 0.997 \\
\bottomrule
\end{tabular}
\caption{Frozen-holdout headline metrics at the five-epoch horizon, richest
feature set, gradient boosting. Reg.\ $R^2$: final-accuracy regression.
Clf.\ AUC: relative-classification ROC-AUC. Fail.\ AUC: failure-prediction
ROC-AUC (a different eligible cohort and range than Clf.\ AUC; see
Section~\ref{sec:protocol}). Majority/mean baselines score
AUC$=0.500$/$R^2\approx0$ in every domain. Complete precision, recall, F1,
RMSE, and per-domain sample sizes for every horizon are given in
Appendix~\ref{app:metrics}.}
\label{tab:headline}
\end{table}

Complete per-horizon metrics for all three endpoints -- final-accuracy
regression (including RMSE), relative classification, and failure
prediction, each across all four horizons and all six domains -- are
given in Appendix~\ref{app:metrics} (Table~\ref{tab:regfull},
Table~\ref{tab:clffull}, Table~\ref{tab:failfull}) and are the basis for
the observation-horizon analysis in Section~\ref{sec:horizonresults}.

\subsection{Prediction by observation horizon}
\label{sec:horizonresults}

\begin{figure}[t]
\centering
\includegraphics[width=\linewidth]{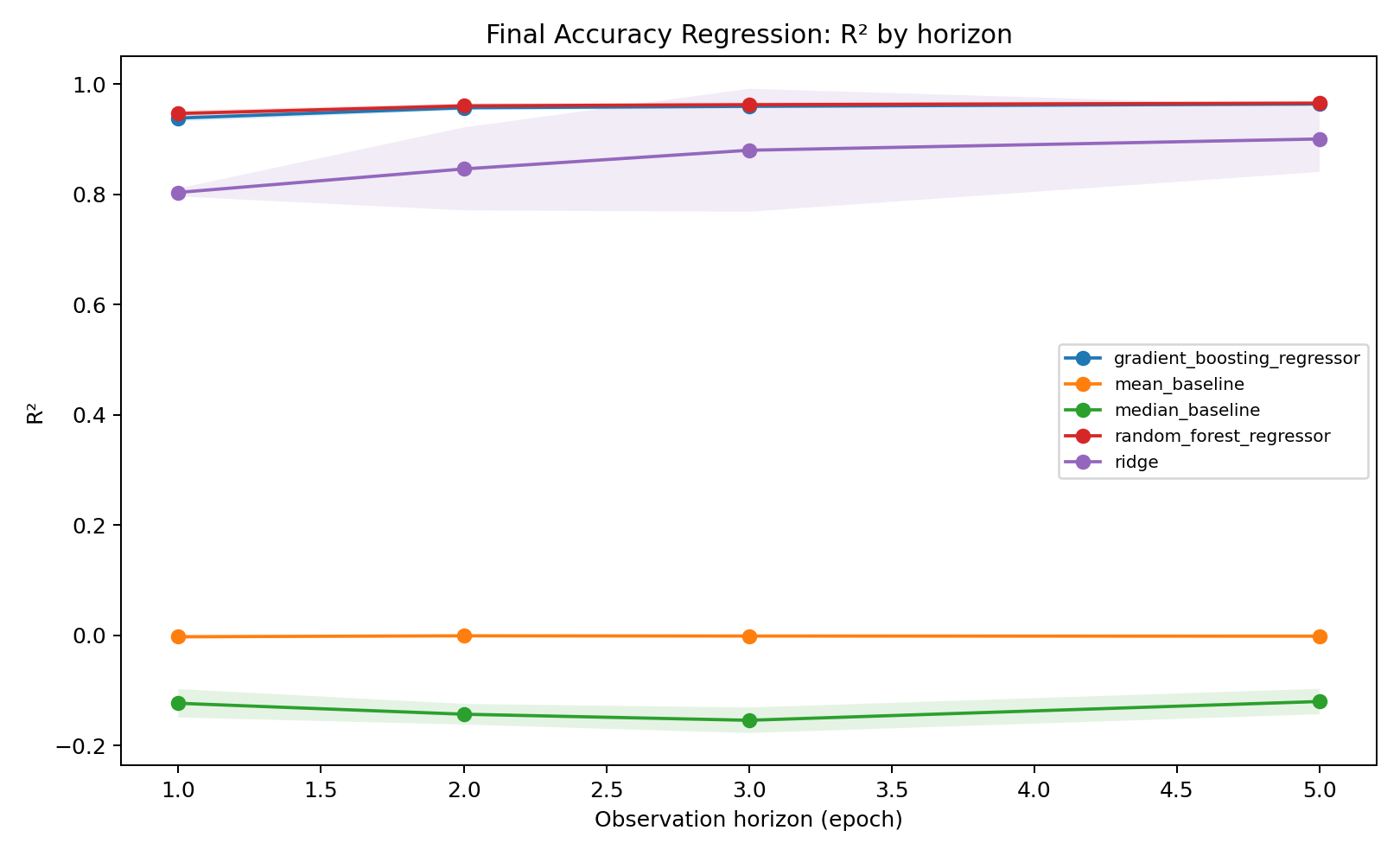}
\caption{Final-accuracy regression $R^2$ by observation horizon
(CIFAR-10/ResNet-18, open 80\% development partition; shaded band is a
Nadeau-Bengio-corrected 95\% confidence interval over 10 resampled splits).
Both tree ensembles are already close to their five-epoch accuracy after a
single epoch, the main evidence behind our claim that useful prediction is
available remarkably early.}
\label{fig:horizon}
\end{figure}

Figure~\ref{fig:horizon} tracks final-accuracy regression $R^2$ across
observation horizons for CIFAR-10/ResNet-18 on the open 80\% development
partition (a different, larger partition than the frozen-holdout headline
above, used here because the horizon curve requires repeated splits at
every horizon rather than a single point estimate). Useful prediction is
available remarkably early: gradient boosting and random forest already
reach $R^2\approx0.94$--$0.95$ after a single epoch, and the gain from
observing four additional epochs is comparatively small for both tree
ensembles. Ridge regression starts substantially further behind
($R^2=0.80$ at $k=1$) and closes roughly half of that gap by $k=5$
($R^2=0.90$), a larger relative improvement than either tree ensemble
shows over the same horizons, though it does not catch up to them. This
indicates that, within the horizons tested, one epoch is already
sufficient to capture most of the linearly-inseparable structure that
longer observation reveals, at least for the non-linear tree-ensemble
models; Appendix~\ref{app:metrics} (Table~\ref{tab:regfull}) shows the
same early-saturation pattern holds for all six domains.

\subsection{Telemetry ablations}
\label{sec:ablation}

\begin{figure}[t]
\centering
\includegraphics[width=0.98\linewidth]{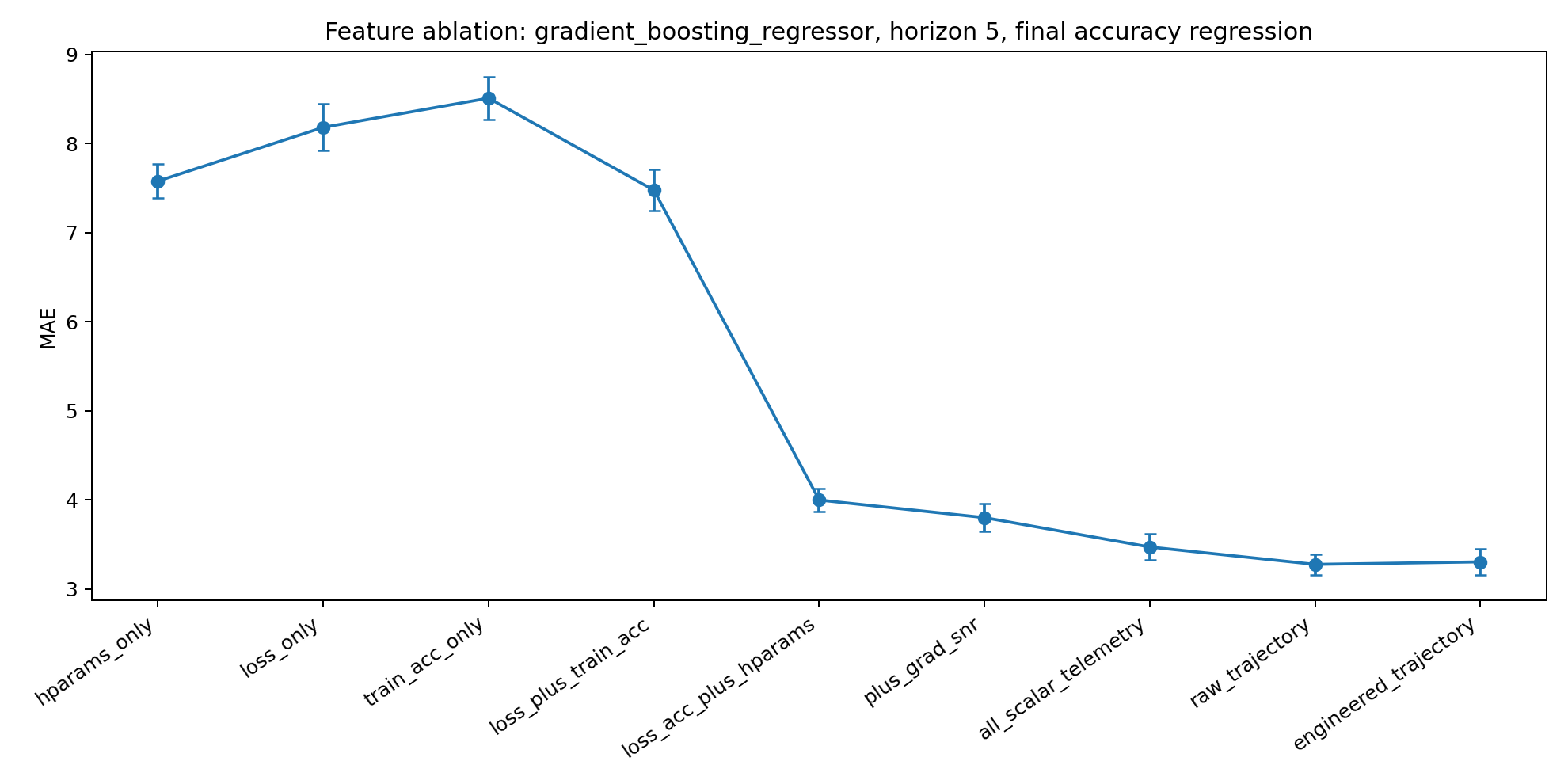}
\caption{Feature-set ablation for final-accuracy regression at the
five-epoch horizon (CIFAR-10/ResNet-18, open 80\%, error bars are
Nadeau-Bengio-corrected 95\% confidence intervals over 10 resampled
splits). The largest error reduction comes from adding hyperparameters, not
internal telemetry; the controlled evidence for telemetry's own
contribution is the curves-only vs.\ curves-plus-internal comparison in
Table~\ref{tab:ablationdeltas}, not this ladder in isolation.}
\label{fig:ablation}
\end{figure}

Figure~\ref{fig:ablation} shows the full feature-set ablation ladder for
CIFAR-10/ResNet-18 at the five-epoch horizon. The single largest reduction
in regression error (MAE $\approx 7.5 \to 4.0$) comes from adding the four
sampled hyperparameters to the loss/accuracy curves, not from internal
telemetry -- unsurprising, since the sampled learning rate and batch size
alone are already informative about whether a configuration will remain
stable. Because hyperparameters by themselves carry this much signal, the
full ladder cannot by itself isolate what internal telemetry specifically
contributes; that requires the paired, hyperparameter-free comparison
described in Section~\ref{sec:featuresets}.

For CIFAR-10/ResNet-18 at the five-epoch horizon, adding gradient
signal-to-noise and weight-norm growth to the curves-only feature set
improves regression $R^2$ by $+0.060$ (95\% CI $[0.052, 0.069]$) and
reduces MAE by $2.62$ (95\% CI $[2.43, 2.81]$). Table~\ref{tab:ablationdeltas}
reports this same paired comparison for all six domains and both
classification tasks. Every one of the eighteen resulting confidence
intervals (six domains $\times$ regression $R^2$ plus twelve
classification-AUC deltas across the two classification tasks) excludes
zero, indicating that internal telemetry provides a statistically
consistent improvement over curves alone everywhere we tested it; the
practical magnitude of that improvement, however, ranges from a
regression-$R^2$ gain as large as $+0.060$ down to as small as $+0.0046$,
and a classification-AUC gain as large as $+0.0206$ down to as small as
$+0.0018$, with the smallest gains (Fashion-MNIST/MLP) too small to matter
in most practical settings even though they are not attributable to chance.

\begin{table*}[t]
\centering
\small
\begin{tabular}{lrr}
\toprule
Domain & $\Delta R^2$ (regression, 95\% CI) & $\Delta$AUC range (both classification tasks) \\
\midrule
C10/ResNet-18 & $+0.0602$ $[0.0518, 0.0685]$ & $[+0.0161, +0.0206]$ \\
C10/SmallCNN & $+0.0332$ $[0.0272, 0.0392]$ & $[+0.0051, +0.0100]$ \\
C10/MLP & $+0.0096$ $[0.0076, 0.0115]$ & $[+0.0018, +0.0042]$ \\
FMNIST/ResNet-18 & $+0.0298$ $[0.0195, 0.0400]$ & $[+0.0089, +0.0104]$ \\
FMNIST/SmallCNN & $+0.0115$ $[0.0043, 0.0187]$ & $[+0.0040, +0.0061]$ \\
FMNIST/MLP & $+0.0046$ $[0.0019, 0.0073]$ & $[+0.0021, +0.0034]$ \\
\bottomrule
\end{tabular}
\caption{Paired, hyperparameter-free improvement from adding gradient
signal-to-noise and weight-norm growth to loss/accuracy curves alone
(curves-plus-internal vs.\ curves-only), five-epoch horizon, open 80\%
development partition, gradient boosting. Every regression interval
excludes zero. $\Delta$AUC range spans the two point-estimate deltas for
the relative-classification and failure-prediction endpoints in that
domain (twelve point estimates total across six domains; every one of the
twelve has an individual 95\% CI, omitted here for brevity, that also
excludes zero).}
\label{tab:ablationdeltas}
\end{table*}

\begin{figure}[t]
\centering
\includegraphics[width=0.98\linewidth]{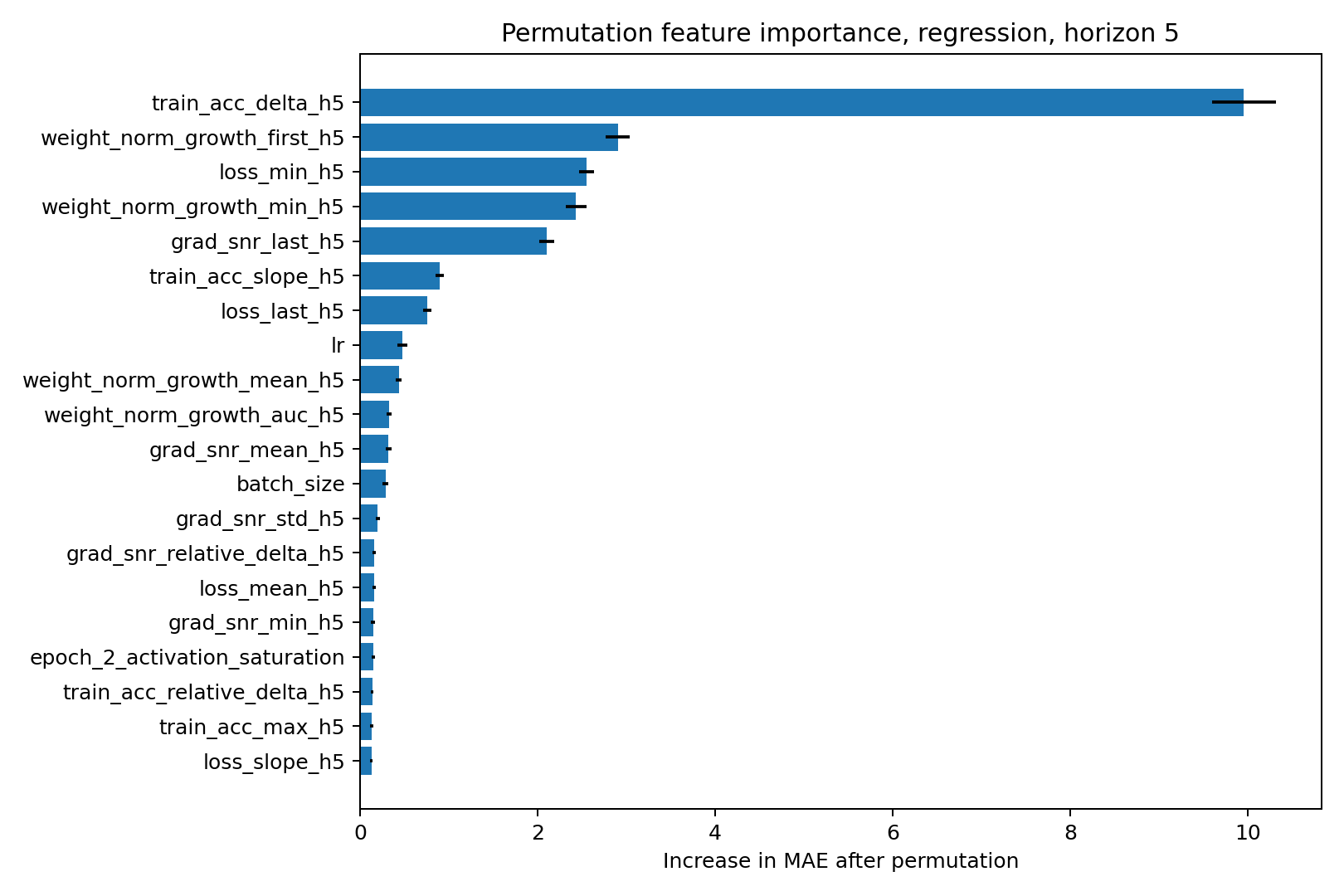}
\caption{Permutation feature importance for final-accuracy regression,
gradient boosting, five-epoch horizon, CIFAR-10/ResNet-18, frozen holdout.
Weight-norm-growth and recent gradient-SNR statistics rank above most raw
loss features, independent evidence that internal telemetry carries
signal beyond what the loss and accuracy curves alone capture.}
\label{fig:importance}
\end{figure}

Permutation feature importance (Figure~\ref{fig:importance}) provides a
second, independent line of evidence: after the change in training
accuracy over the horizon, which is by far the single strongest feature,
weight-norm-growth statistics and the most recent gradient
signal-to-noise reading outrank most of the raw loss statistics, while
activation saturation -- measured only once rather than tracked across the
horizon like the other three signals -- ranks near the bottom. This
appears, at first glance, to contradict Figure~\ref{fig:ablation}, where
adding hyperparameters produces the single largest drop in error, yet the
sampled learning rate and batch size themselves rank low in permutation
importance. We believe this is mainly due to feature collinearity rather than a
genuine contradiction: once the engineered trajectory features are already
in the model, much of what the hyperparameters would otherwise convey is
already recoverable from how the trajectory itself behaves, so permuting
the hyperparameters alone does not destroy much additional predictive
information that is not also carried
elsewhere in the feature set.

\subsection{Cross-architecture transfer}
\label{sec:crossarch}

\begin{figure}[t]
\centering
\includegraphics[width=0.98\linewidth]{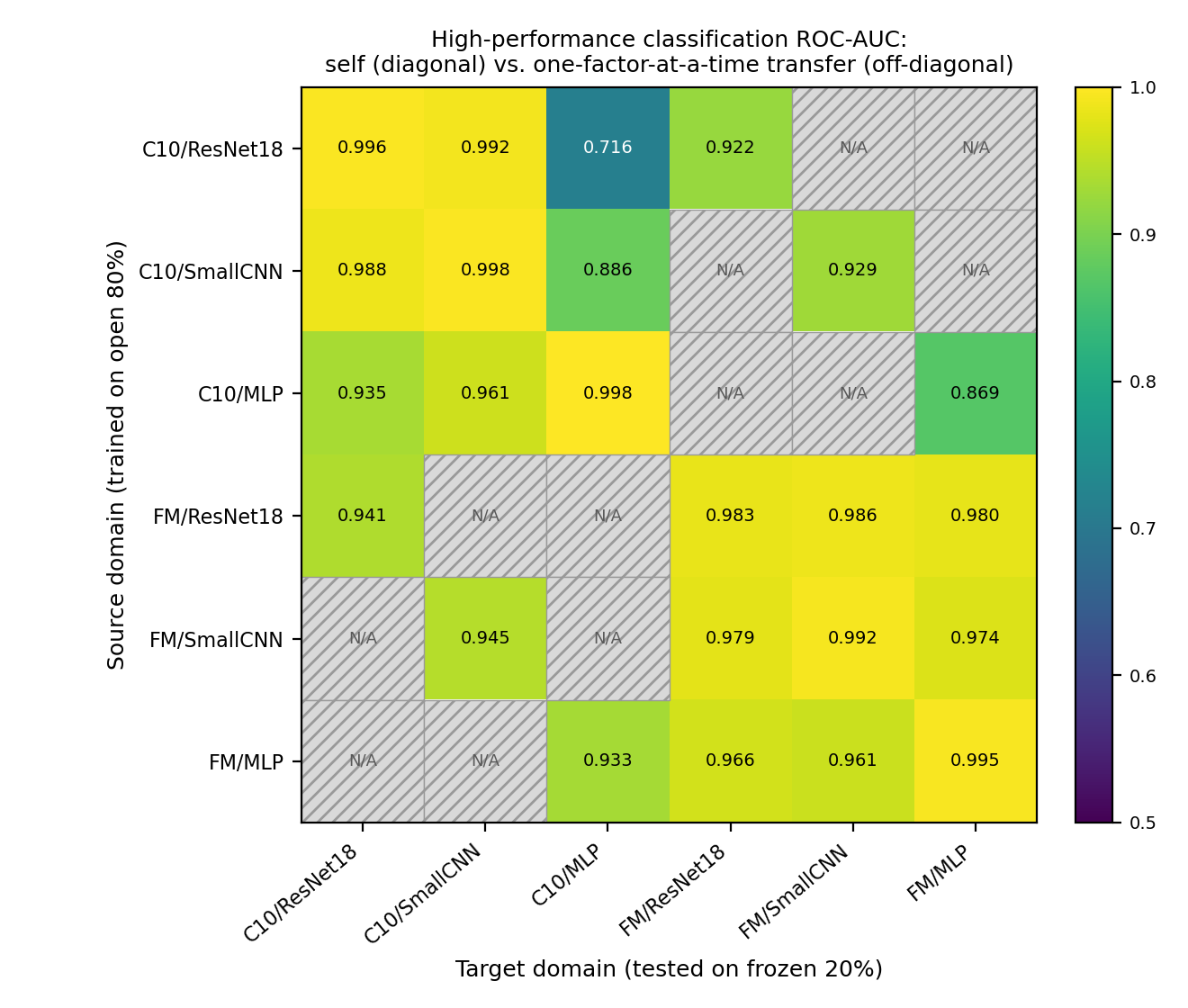}
\caption{Relative-classification ROC-AUC for a predictor trained on the
source domain's open 80\% partition and evaluated, with no retraining, on
the target domain's frozen 20\% partition. Diagonal cells are in-domain
scores. Off-diagonal cells are shown only for pairs differing in exactly
one factor (architecture with dataset fixed, or dataset with architecture
fixed); hatched gray cells differ in both simultaneously and are reported
separately in Appendix~\ref{app:transfer} so the two transfer effects are
not conflated. Transfer is strong almost everywhere except one direction
into the multilayer perceptron on CIFAR-10 (top row, third column), the
one clear outlier in the matrix.}
\label{fig:transfer}
\end{figure}

To separate the effects of changing architecture and changing dataset, we
evaluate every ordered pair among our six domains that differs in exactly
one of \{architecture, dataset\}: eighteen pairs in total, shown in
Figure~\ref{fig:transfer} together with the six in-domain scores on the
diagonal. Cross-architecture transfer between the two convolutional
architectures is strong in both directions on both datasets:
ResNet-18$\leftrightarrow$SmallCNN reaches classification ROC-AUC
$\geq 0.979$ and Spearman rank correlation $\geq 0.89$ regardless of
direction or dataset.

Transfer involving the multilayer perceptron is more dataset-dependent. On
CIFAR-10, transferring a predictor from ResNet-18 to the multilayer
perceptron is a clear outlier (Spearman $0.604$, AUC $0.716$), well below
every other one-factor pair we measured; the analogous transfer from
SmallCNN (Spearman $0.913$, AUC $0.886$) is also below most other pairs,
though far less extreme. Transfer in the opposite direction -- from the
multilayer perceptron into either convolutional architecture -- remains
strong on CIFAR-10 (MLP$\to$ResNet-18: AUC $0.935$; MLP$\to$SmallCNN: AUC
$0.961$), so on this dataset a predictor trained on the multilayer
perceptron's telemetry generalizes to convolutional architectures better
than the reverse. This asymmetry does not hold on Fashion-MNIST, where
transfer is strong in every direction (AUC $0.961$--$0.986$) and, if
anything, the direction of the CIFAR-10 asymmetry is reversed
(ResNet-18$\to$MLP: AUC $0.980$ vs.\ MLP$\to$ResNet-18: AUC $0.966$ on
Fashion-MNIST). Both the CIFAR-10 outlier and its direction therefore
appear specific to that dataset rather than a general property of the
multilayer perceptron architecture (Section~\ref{sec:discussion}).

\subsection{Cross-dataset transfer}
\label{sec:crossdataset}

Cross-dataset transfer (holding architecture fixed) exhibits a large
scale-mismatch effect in raw regression error:
CIFAR-10/ResNet-18$\to$Fashion-MNIST/ResNet-18 transfer achieves MAE
$=31.3$, roughly eight times that domain's own in-domain MAE of $3.8$. This
is broadly consistent with, though somewhat larger than, the 21.8-point
gap between the two datasets' typical final accuracies (medians $66.8$ vs.\
$88.6$), suggesting that scale mismatch is the primary but not the only
driver of the inflated raw error. Despite this, Spearman rank correlation
under transfer remains moderate to high across all six cross-dataset pairs
($0.52$--$0.89$), and relative-classification ROC-AUC under
domain-relative labeling remains high throughout ($0.869$--$0.945$, lowest
for CIFAR-10/MLP$\to$Fashion-MNIST/MLP). The \emph{shape} of the
relationship between early telemetry and relative performance therefore
transfers across datasets considerably better than its absolute numerical
scale does, which is why we report both raw error and rank-based/relative
metrics throughout this evaluation rather than raw error alone.

The twelve additional pairs that differ in both architecture and dataset
simultaneously (Appendix~\ref{app:transfer}) show no simple, additive
degradation from combining both factors: some ``both-differ'' pairs score
below both of their corresponding one-factor comparators, while others --
for instance, CIFAR-10/ResNet-18$\to$Fashion-MNIST/SmallCNN, which differs
in both factors, scores higher (AUC $0.938$) than the matched
single-factor cross-dataset pair CIFAR-10/ResNet-18$\to$Fashion-MNIST/ResNet-18
(AUC $0.922$). Given that even the clean one-factor categories already
span a wide range on their own (CIFAR-10/ResNet-18's cross-architecture
scores alone range from $0.716$ to $0.992$ depending on target), which
single factor dominates transfer difficulty appears to depend on the
specific source/target pair rather than following a simple additive rule.

\subsection{Error analysis}
\label{sec:erroranalysis}

\begin{figure*}[t]
\centering
\includegraphics[width=0.98\linewidth]{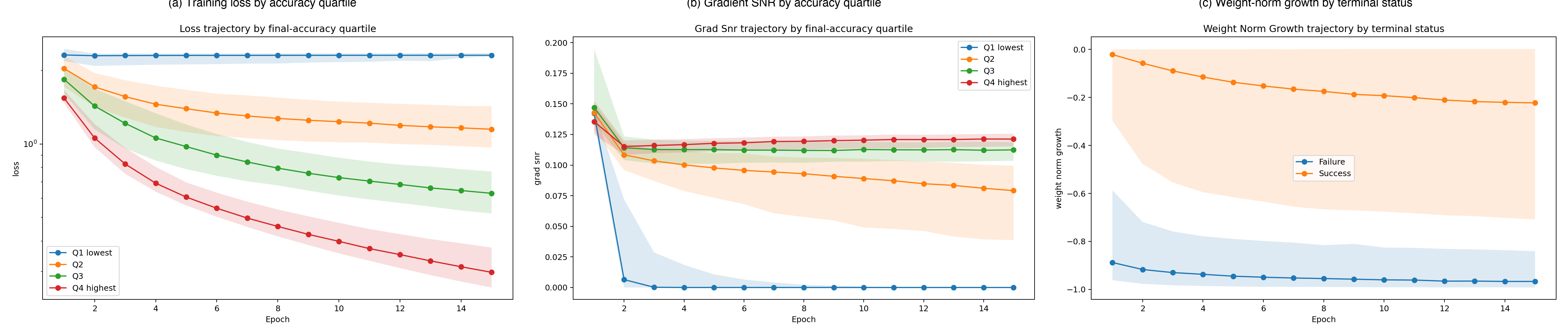}
\caption{Three telemetry signals decomposed by outcome for
CIFAR-10/ResNet-18: (a) training loss by final-accuracy quartile, (b)
gradient signal-to-noise ratio by final-accuracy quartile, (c) weight-norm
growth by terminal status (failure merges the below-threshold and
numerical-divergence outcomes into one group against success). Shaded
bands are one standard deviation across runs in each group. Outcome groups
are visibly separated within the first few epochs in every panel, the main
qualitative evidence behind the strong early-horizon results in
Section~\ref{sec:horizonresults}.}
\label{fig:trajectories}
\end{figure*}

Figure~\ref{fig:trajectories} decomposes three of our telemetry signals by
outcome to motivate which configurations are hardest to predict correctly.
Higher-accuracy runs separate from lower-accuracy runs in their training
loss within the first few epochs (panel a), which is consistent with why
useful regression is already possible at the one-epoch horizon
(Section~\ref{sec:horizonresults}); gradient signal-to-noise separates the
lowest-accuracy quartile sharply from the rest almost immediately, but
distinguishes less clearly among the top three quartiles (panel b); and
weight-norm growth separates eventual failures from eventual successes by
around epoch 2--3 in most cases, though a diverged run's weight norm
remains plottable for every epoch it completed before diverging (panel c).

\begin{figure}[t]
\centering
\includegraphics[width=\linewidth]{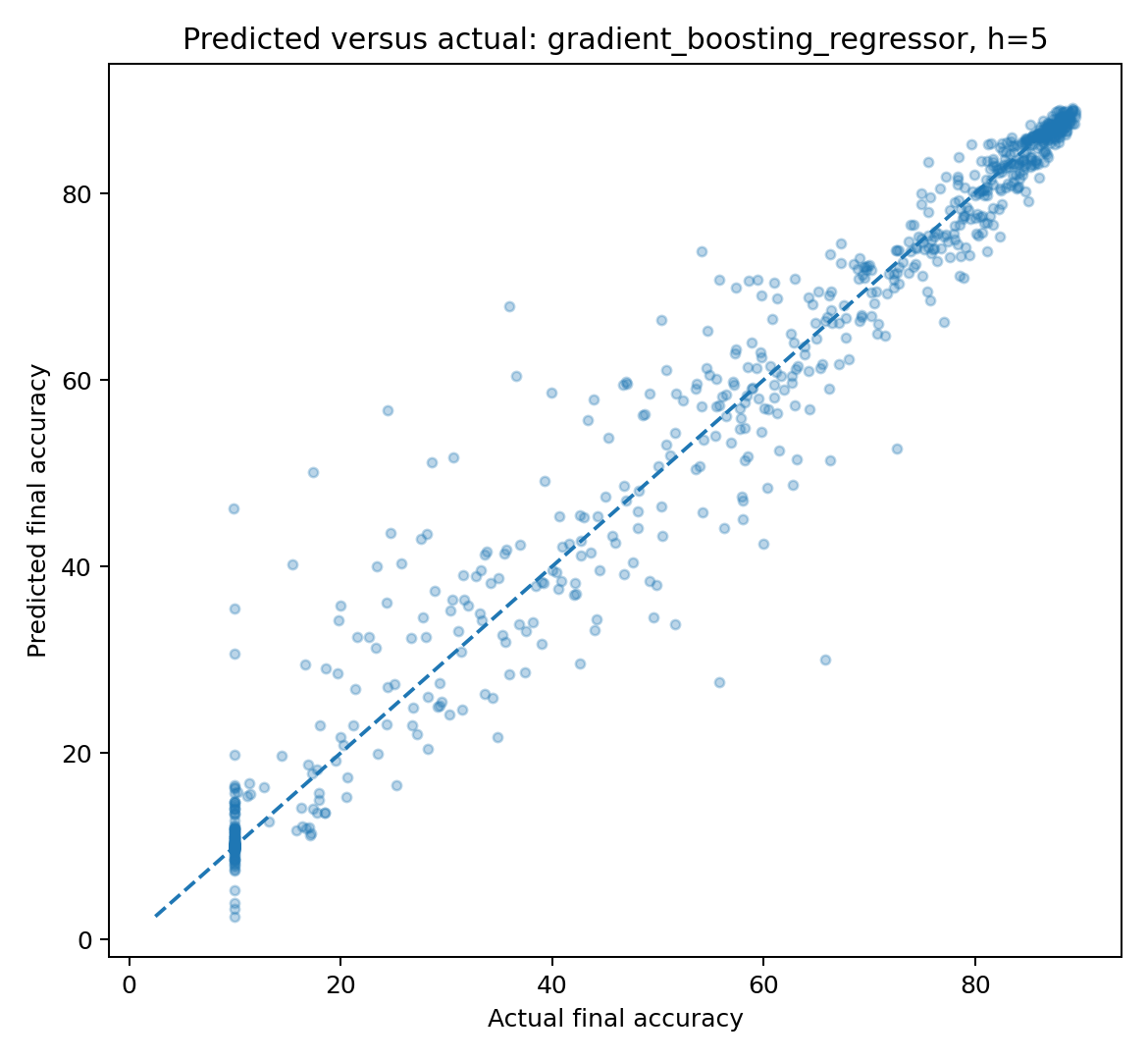}
\caption{Predicted vs.\ actual final accuracy, gradient boosting,
five-epoch horizon, CIFAR-10/ResNet-18, frozen holdout. The cluster of
predictions between roughly 0 and 45 for runs whose actual accuracy is
near chance is the boundary-region failure case discussed above: a
moderate-looking early trajectory that collapses only after the
observation horizon ends.}
\label{fig:predvact}
\end{figure}

The ten largest regression errors, and most of the failure classifier's 19
false negatives (out of 289 later-onset failures in the frozen holdout;
recall $=0.934$), concentrate on an overlapping set of configurations:
runs whose first five epochs resemble a reasonable, moderate learning
trajectory but that ultimately either plateau just under the domain's
usability threshold or collapse toward chance accuracy only after the
observation horizon has ended. Figure~\ref{fig:predvact} shows this
directly as a cluster of predictions between roughly 0 and 45 sitting well
above the actual outcome for runs whose true final accuracy is near chance
level ($\approx10\%$ for 10-class CIFAR-10). Because the second sampling
phase deliberately oversamples the region immediately around the estimated
success/failure boundary (Section~\ref{sec:sampling}), this is precisely
the population the study was designed to probe, and it is also precisely
the population for which five epochs of observation are least sufficient to
distinguish a configuration that will narrowly clear the usability
threshold from one that will plateau just under it or collapse later.

\section{Discussion}
\label{sec:discussion}

\textbf{Why prediction is already strong after one epoch.} The
trajectory-by-quartile decomposition in Figure~\ref{fig:trajectories}
offers a direct explanation: the loss curves of eventually high- and
low-accuracy runs are already visibly separated within the very first
epoch, particularly at the extremes. Coupled with the observation
(Section~\ref{sec:limitations}) that some frozen-holdout configurations
plausibly sit close in hyperparameter space to open-partition
configurations under our boundary-focused sampling design, we think the
strong one-epoch result reflects two things at once: a genuine early
signal in the training dynamics, and an evaluation design that likely
makes the prediction problem somewhat easier than it would be under fully
independent sampling of the frozen partition. We cannot fully separate
these two contributions with the current data.

\textbf{Hyperparameters versus internal telemetry.} The two are not
redundant because they answer different questions. Sampled hyperparameters
are a property of the search space itself, knowable in principle before
training starts; internal telemetry instead measures what is actually
happening in one specific realized trajectory. That distinction is why
telemetry still helps even after hyperparameters are already in the
feature set (Table~\ref{tab:ablationdeltas}): a hyperparameter combination
that is usually stable can still produce an unstable run, and telemetry is
what reveals that a particular run is the exception.

\textbf{Statistically consistent but practically small effects.}
Statistical significance under a specific resampling protocol is a
different claim from practical usefulness. The smallest telemetry gains we
observe cluster in domains where curves and hyperparameters alone already
achieve very high accuracy (Table~\ref{tab:headline}); once a feature set
is already close to a domain's accuracy ceiling, there is simply less
remaining signal left for any additional feature to capture, telemetry
included.

\textbf{Scale mismatch versus relationship transfer.} Raw regression error
alone is a misleading way to evaluate cross-dataset transfer, because
absolute error inflates whenever source and target domains occupy
different accuracy ranges even when the underlying relationship between
telemetry and relative outcome is preserved almost intact. Reporting rank
correlation and domain-relative classification alongside raw error is what
makes this distinction visible: raw error alone would overstate the
failure of transfer, and relative metrics alone would understate the
practical cost of applying a source model's raw predictions unmodified to
a new domain.

\textbf{The CIFAR-10 architecture asymmetry.} We do not have a confirmed
mechanistic explanation for why transferring specifically \emph{into} the
multilayer perceptron from a convolutional architecture on CIFAR-10 is a
clear outlier while every other direction, dataset, and pairing transfers
comparatively well. Two observations bound the space of plausible
explanations without settling on one: the asymmetry is direction-specific
(transfer \emph{out of} the multilayer perceptron remains strong on the
same dataset), and it is dataset-specific (the same architecture pair
transfers well in both directions on Fashion-MNIST). This is consistent
with an interaction between the multilayer perceptron's much lower
achievable accuracy ceiling on CIFAR-10 specifically (Table~\ref{tab:headline})
and whatever telemetry pattern a convolutional source model has learned to
associate with success, but we present this as a hypothesis for future
investigation, not a finding our current experiments confirm directly.

\textbf{Boundary-focused sampling and interpretation.} Every error rate,
class balance, and confidence interval reported in this paper describes a
population deliberately enriched near the success/failure boundary
(Section~\ref{sec:sampling}). This is a deliberate design choice that makes
the prediction problem harder and more informative than an arbitrary
sweep's outcome distribution would be, but it also means our reported
metrics should not be read as an estimate of how well this approach would
perform on a differently-shaped, less boundary-heavy population; a
practitioner's typical sweep, with fewer configurations near the boundary,
would likely see different absolute error rates even under an identical
modeling approach.

\textbf{Decision support, not automatic termination.} We frame our results
throughout as evidence for a decision-support signal, not a basis for
automatically terminating training runs without oversight: a confident
early failure prediction is still a prediction, subject to the false
negatives documented in Section~\ref{sec:erroranalysis}. Section~\ref{sec:ethics}
expands on this distinction.

\section{Limitations}
\label{sec:limitations}

We summarize the conditions under which the above findings should be
interpreted, several of which are noted where relevant above.

\textbf{Scope of architectures and datasets.} All results are drawn from
three architecture families on two image classification datasets. We do
not study large language models, large-scale vision transformers or
convolutional networks at modern scale, or non-vision domains; whether
early-training telemetry carries comparable predictive signal in those
settings is untested.

\textbf{Short training schedules.} Every run is capped at 15 epochs. This
is sufficient to observe both successful and failing training dynamics in
our setting, but it does not establish that the same telemetry signals, or
the same horizon-1 saturation pattern (Section~\ref{sec:horizonresults}),
would hold for training schedules an order of magnitude longer; whether
prediction remains possible, and from how early, for much longer schedules
is a natural question this study does not answer and we leave to future
work.

\textbf{Limited hyperparameter and optimizer diversity.} We vary four
hyperparameters (learning rate, batch size, weight decay, batch
normalization use) and do not vary the optimizer family or learning-rate
schedule beyond what is implicitly covered by these four axes. Predictive
signal derived from a different or more diverse hyperparameter space, or a
different optimizer, is untested.

\textbf{Frozen-holdout proximity in hyperparameter space.} As noted in
Section~\ref{sec:frozenholdout}, because our boundary-focused sampling
concentrates draws in a small region of a four-dimensional hyperparameter
space, some frozen-holdout configurations likely lie close to, though never
identical to, an open-partition configuration used during development.
This is a partial, unconfirmed explanation for part of the strong
early-horizon result and should temper a literal reading of the frozen
holdout as a test of generalization to an arbitrarily distant new
configuration.

\textbf{No cross-seed repeats.} Our data contains no repeated training runs
of an identical hyperparameter configuration under different random seeds.
The grouped-holdout protocol (Section~\ref{sec:evalsplits}) is designed to
protect against such repeats splitting across train and test, but with zero
repeats actually present in the data, this protection was not exercised in
practice, and we cannot separately estimate seed-to-seed variance at fixed
hyperparameters.

\textbf{Single-domain neural predictor.} The feed-forward neural network
predictor (Section~\ref{sec:indomain}) was trained and evaluated in only
one of six domains, due to the added cost of training and tuning a
separate neural predictor per domain; we cannot confirm that its rough
parity with gradient boosting generalizes to the other five domains.

\textbf{Point estimates from a single frozen partition.} Every
frozen-holdout headline number is a point estimate from one train/test
partition, not an average over repeated frozen splits, since the frozen
partition is by design used exactly once. Small, non-monotonic
fluctuations across observation horizons in the complete per-horizon
tables (Appendix~\ref{app:metrics}) should be read as ordinary
single-split estimation noise rather than as evidence that additional
telemetry ever degrades prediction.

\textbf{Manually chosen, domain-specific usability thresholds.} The
threshold separating a usable from an unusable final accuracy in each
domain (Table~\ref{tab:accounting}) was chosen manually, before data
generation, using domain knowledge about achievable accuracy on each
dataset/architecture combination. A different, defensible choice of
threshold would shift the failure-prediction task's class balance and
could shift its reported metrics.

\textbf{Strong early prediction is not proof of safe automatic stopping.}
As discussed in Section~\ref{sec:discussion} and
Section~\ref{sec:ethics}, high predictive accuracy on this evaluation does
not by itself establish that automatically terminating runs on the basis
of these predictions, without human oversight, would be a safe or
desirable practice.

\section{Ethical and Practical Considerations}
\label{sec:ethics}

The intended use of a system like the one studied here is to help a human
operator allocate limited compute across a large hyperparameter sweep --
for example, by surfacing which currently running configurations are
predicted to be at high risk of failure or of underperforming, so that a
person can decide whether to reallocate that compute elsewhere. We do not
propose or endorse using these predictions to terminate training runs
fully automatically.

\textbf{Risk of terminating genuinely good runs.} A false-positive failure
prediction that leads to automatic termination forecloses the possibility
that the run recovers or that the prediction was simply wrong
(Section~\ref{sec:erroranalysis} documents a non-trivial false-negative
rate even at our best-performing horizon and domain, and every classifier
in this study has some false-positive rate as well). A human-in-the-loop
deployment allows an incorrect early prediction to be caught and
overridden before it has an irreversible effect on a specific run; a fully
automated one does not.

\textbf{Compute efficiency and environmental cost.} Training deep networks
to convergence has a real energy and compute cost, and a large fraction of
sweep compute is spent on runs that were foreseeably going to fail. The
telemetry we use is cheap to collect -- a handful of scalar statistics
computed from a training process that is already running -- so even a
modest, reliable early-warning signal could reduce wasted compute, and the
associated energy use, without requiring a more accurate or more expensive
predictor. We see this as a complement to, not a replacement for, existing
comparison-based scheduling methods (Section~\ref{sec:related}), which
already capture some of this benefit.

\textbf{Population shift.} Our reported error rates and class balances are
specific to an evaluation cohort deliberately enriched near the
success/failure boundary (Section~\ref{sec:sampling}); anyone deploying a
similar predictor on a differently-shaped population of configurations
should re-validate its error rates on that population rather than assume
our reported numbers transfer directly.

\textbf{Human oversight.} Every result in this paper is intended as
decision support for a person managing a sweep, not as an autonomous
control system. Given the false-negative and false-positive rates
documented throughout, and given that boundary-region runs
(Section~\ref{sec:erroranalysis}) are simultaneously the hardest to
predict and the most common in our enriched evaluation cohort, we think
human review remains necessary before any run is terminated on the basis
of a prediction like ours.

\section{Conclusion}
\label{sec:conclusion}

We studied whether a single training run's own early telemetry, in
isolation from any other run, is sufficient to predict its final accuracy,
its relative standing within its cohort, and its risk of an outright
training-dynamics failure. Across 23{,}788 training runs spanning six
architecture/dataset combinations, gradient-boosted trees using only the
first few epochs of loss, accuracy, gradient, and weight-norm telemetry,
together with the run's sampled hyperparameters, achieve strong prediction
on a permanently held-out partition of configurations, with most of that
predictive power already available after a single epoch. A controlled,
hyperparameter-free ablation shows that gradient- and weight-level internal
telemetry specifically, not just the loss and accuracy curves every
practitioner already watches, provides a statistically consistent
improvement in every domain and task we tested, even though the practical
size of that improvement is domain-dependent and sometimes small. Transfer
across architecturally similar domains is strong; transfer across datasets
is limited mainly by a mismatch in absolute accuracy scale rather than a
loss of the underlying relationship between telemetry and relative
performance, and one specific architecture pairing on one dataset
transfers noticeably worse than every other combination we measured, for
reasons we can characterize but not yet fully explain.

Several extensions follow directly from the scope of this study
(Section~\ref{sec:limitations}). Longer training schedules would show
whether an early prediction continues to hold up or gets revised as more
of a run's trajectory becomes visible. Additional architecture families,
datasets beyond image classification, and a wider hyperparameter space
that also varies the optimizer and learning-rate schedule would test how
much of what we observe here is specific to our setting. Extending the
feed-forward neural predictor beyond the single domain evaluated in this
study, and evaluating telemetry-based prediction under sampling that is
not deliberately enriched near a success/failure boundary, would both
speak directly to how these results generalize to a practitioner's
ordinary sweep rather than our boundary-focused evaluation cohort.

More broadly, we read these results as evidence that the internal state of
an optimization process -- not just the loss and accuracy values it
produces -- carries information about where that process is headed, and
that this information is cheap enough to collect that it is worth treating
as a routine part of training telemetry rather than a specialized
diagnostic. As hyperparameter sweeps and large-scale training runs
continue to grow in number and cost, even a modest, well-calibrated
early-warning signal, used as decision support alongside a human operator
rather than as an autonomous control system, could meaningfully change how
that compute is spent.

{\small
\bibliographystyle{ieee_fullname}
\bibliography{egbib}
}

\clearpage
\onecolumn

\appendix

\section{Detailed Experimental Setup}
\label{app:setup}

\textbf{Hyperparameter sampling ranges.} All four sampled hyperparameters
are drawn from fixed ranges shared across the two sampling phases described
in Section~\ref{sec:sampling}; the second phase reweights draws toward the
estimated success/failure boundary within these same ranges rather than
sampling from a different range. Learning rate and weight decay are sampled
on a logarithmic scale; learning rate spans several orders of magnitude
from a small stable value up to a value large enough to reliably destabilize
training for every architecture in this study, and weight decay spans from
effectively zero regularization up to a strength large enough to
substantially suppress learning. Batch size is sampled from a small
discrete set of powers of two. Batch normalization use is sampled as a
binary indicator. The exact numeric endpoints of each range were tuned
per-architecture during pilot runs so that both the ``obviously stable''
and ``obviously unstable'' regions of the search space are populated for
every architecture, which is a prerequisite for the boundary-focused
second sampling phase to have a meaningful boundary to concentrate around.

\textbf{Two-phase sampling, restated in full.} The first phase draws all
four hyperparameters independently and broadly for a fixed number of runs
per domain, producing the initial population from which
Table~\ref{tab:status} (Appendix~\ref{app:dataaccounting}) reports terminal
outcomes. We then fit a lightweight probabilistic classifier -- logistic
in the log-transformed hyperparameters -- to these first-phase outcomes,
predicting the probability that a given hyperparameter configuration
results in a training-dynamics failure. The second phase draws
approximately half its configurations by rejection sampling against this
fitted failure-probability model, accepting configurations whose estimated
failure probability is close to one half (i.e., near the estimated
decision boundary) more often than configurations confidently predicted to
succeed or fail, and draws the remaining approximately half of its
configurations using the same broad, unweighted sampling as the first
phase, to keep the domain's overall hyperparameter space populated outside
the boundary region as well. This mixture is what produces the
boundary-enriched but not boundary-only population analyzed throughout the
paper.

\textbf{Training details.} Every run trains its architecture from
random initialization for exactly 15 epochs (or until numerical
divergence, if earlier) using stochastic gradient descent with
cross-entropy loss, with the sampled learning rate, batch size, weight
decay, and batch-normalization setting applied throughout training; no
learning-rate schedule, data augmentation policy, or other hyperparameter
is varied across runs beyond these four.

\textbf{Compute.} All training runs were executed on GPU-accelerated
compute nodes managed by a Slurm-based cluster scheduler; specific cluster
identifiers, job-array configurations, and account details are omitted as
not scientifically relevant to reproducing the study's methodology.

\section{Telemetry Definitions and Feature Construction}
\label{app:telemetry}

\textbf{Gradient signal-to-noise ratio.} Before training begins, we draw a
fixed random subset of up to 200{,}000 parameter coordinates once per run,
sampled proportionally across parameter tensors so the statistic is not
biased toward whichever layers happen to be enumerated first. During epoch
$e$, for each of the first $\min(40, \text{batches per epoch})$
mini-batches, we record the gradient value at each sampled coordinate
immediately after backpropagation and before the optimizer update is
applied. Writing $g_{t,i}$ for the recorded gradient at coordinate $i$ on
mini-batch $t$ within the epoch, and $I$ for the sampled coordinate set,
\[
\text{grad\_snr}_e = \frac{1}{|I|}\sum_{i\in I}
\frac{\left|\,\text{mean}_t(g_{t,i})\,\right|}{\text{std}_t(g_{t,i}) + 10^{-8}}.
\]
A high value indicates that the gradient at a coordinate points in a
consistent direction across mini-batches within the epoch; a low value
indicates a noisy, inconsistent direction.

\textbf{Weight-norm growth.} Let $\theta_0$ denote the model's parameters
immediately after initialization, before any training step, and $\theta_e$
the parameters at the end of epoch $e$. Then
\[
\text{weight\_norm\_growth}_e = \frac{\lVert\theta_e\rVert_2 - \lVert\theta_0\rVert_2}{\lVert\theta_0\rVert_2 + 10^{-8}},
\]
the relative change, since initialization, in the $L_2$ norm of all
trainable parameters summed across every parameter tensor in the network.

\textbf{Activation saturation.} Measured once, at a fixed early epoch
(epoch 2 by default). Over three batches drawn from the training data, the
model is run in evaluation mode and we record the fraction of elements at
or below zero in the penultimate-layer activation (or, for architectures
without a well-defined penultimate representation, the full model output):
\[
\text{activation\_saturation} = \frac{\#\{\text{activations} \le 0\}}{\#\{\text{activations}\} + 10^{-8}}.
\]
Because this is a single-epoch snapshot rather than a per-epoch trajectory
like the other three signals, it is included in the feature set only from
the observation horizon at or after its measurement epoch onward, and
excluded entirely at any earlier horizon (Section~\ref{sec:telemetry}).

\textbf{Engineered trajectory features.} For each of the four per-epoch
telemetry signals (loss, training accuracy, gradient signal-to-noise,
weight-norm growth), and at each observation horizon $k$, our richest
feature set (Section~\ref{sec:featuresets}) computes ten summary
statistics over the signal's values through epoch $k$: the first value,
the last value, the change from first to last, the mean, the standard
deviation, the minimum, the maximum, the area under the curve (trapezoidal
approximation over the observed epochs), the slope of a linear fit over
the observed epochs, and the number of sign reversals in the epoch-to-epoch
change. Heavy-tailed raw quantities (loss, weight-norm growth, gradient
signal-to-noise) are log- or signed-log-transformed before being used by
linear or distance-based models; tree-ensemble models use the untransformed
values directly, since tree splits are invariant to monotonic
transformations of a single feature.

\section{Complete Per-Domain and Per-Horizon Metrics}
\label{app:metrics}

Table~\ref{tab:regfull} gives complete final-accuracy regression metrics
across all four observation horizons and all six domains, extending the
horizon-5 summary in Table~\ref{tab:headline} and underlying the
observation-horizon analysis in Section~\ref{sec:horizonresults}.

\begin{table*}[t]
\centering
\small
\begin{tabular}{lrrrrr}
\toprule
Domain & $k$ & MAE & RMSE & $R^2$ & Spearman \\
\midrule
C10/ResNet-18 & 1 & 4.54 & 7.30 & 0.941 & 0.970 \\
C10/ResNet-18 & 2 & 3.51 & 6.23 & 0.957 & 0.979 \\
C10/ResNet-18 & 3 & 3.51 & 6.06 & 0.959 & 0.980 \\
C10/ResNet-18 & 5 & 3.23 & 5.69 & 0.964 & 0.981 \\
C10/SmallCNN & 1 & 5.92 & 9.87 & 0.899 & 0.947 \\
C10/SmallCNN & 2 & 4.11 & 6.99 & 0.949 & 0.969 \\
C10/SmallCNN & 3 & 3.81 & 6.54 & 0.956 & 0.971 \\
C10/SmallCNN & 5 & 3.47 & 6.53 & 0.956 & 0.975 \\
C10/MLP & 1 & 1.60 & 2.60 & 0.974 & 0.983 \\
C10/MLP & 2 & 1.28 & 2.02 & 0.984 & 0.987 \\
C10/MLP & 3 & 1.17 & 1.89 & 0.986 & 0.989 \\
C10/MLP & 5 & 0.95 & 1.57 & 0.990 & 0.991 \\
FMNIST/ResNet-18 & 1 & 6.56 & 12.17 & 0.844 & 0.919 \\
FMNIST/ResNet-18 & 2 & 4.66 & 9.57 & 0.903 & 0.950 \\
FMNIST/ResNet-18 & 3 & 4.06 & 8.97 & 0.915 & 0.957 \\
FMNIST/ResNet-18 & 5 & 3.83 & 8.57 & 0.923 & 0.962 \\
FMNIST/SmallCNN & 1 & 6.07 & 12.10 & 0.877 & 0.882 \\
FMNIST/SmallCNN & 2 & 4.06 & 9.25 & 0.928 & 0.956 \\
FMNIST/SmallCNN & 3 & 3.33 & 8.71 & 0.936 & 0.961 \\
FMNIST/SmallCNN & 5 & 2.89 & 7.86 & 0.948 & 0.965 \\
FMNIST/MLP & 1 & 3.38 & 7.52 & 0.915 & 0.940 \\
FMNIST/MLP & 2 & 2.11 & 4.49 & 0.970 & 0.971 \\
FMNIST/MLP & 3 & 2.03 & 4.86 & 0.965 & 0.976 \\
FMNIST/MLP & 5 & 1.61 & 3.45 & 0.982 & 0.982 \\
\bottomrule
\end{tabular}
\caption{Final-accuracy regression, all four observation horizons and all
six domains, frozen holdout, richest feature set, gradient boosting.}
\label{tab:regfull}
\end{table*}

Table~\ref{tab:clffull} and Table~\ref{tab:failfull} give complete
per-horizon metrics for relative classification and failure prediction,
respectively, across all six domains, extending the horizon-5 summary in
Table~\ref{tab:headline}. $n$ is the eligible cohort size at that horizon
and domain (Section~\ref{sec:protocol}); ``Prev.'' is the prevalence of the
positive class.

\begin{table*}[t]
\centering
\scriptsize
\begin{tabular}{lrrrrrrrrr}
\toprule
Domain & $k$ & $n$ & Prev. & Acc. & Bal.Acc. & Prec. & Rec. & F1 & AUC \\
\midrule
C10/ResNet-18 & 1 & 778 & 0.47 & 0.946 & 0.947 & 0.924 & 0.964 & 0.944 & 0.992 \\
C10/ResNet-18 & 2 & 778 & 0.47 & 0.960 & 0.961 & 0.937 & 0.981 & 0.959 & 0.995 \\
C10/ResNet-18 & 3 & 778 & 0.47 & 0.961 & 0.962 & 0.947 & 0.973 & 0.960 & 0.996 \\
C10/ResNet-18 & 5 & 778 & 0.47 & 0.959 & 0.960 & 0.937 & 0.978 & 0.957 & 0.996 \\
C10/SmallCNN & 1 & 754 & 0.51 & 0.969 & 0.970 & 0.971 & 0.969 & 0.970 & 0.997 \\
C10/SmallCNN & 2 & 754 & 0.51 & 0.966 & 0.965 & 0.959 & 0.974 & 0.966 & 0.996 \\
C10/SmallCNN & 3 & 754 & 0.51 & 0.971 & 0.971 & 0.966 & 0.977 & 0.971 & 0.997 \\
C10/SmallCNN & 5 & 754 & 0.51 & 0.980 & 0.980 & 0.977 & 0.984 & 0.980 & 0.998 \\
C10/MLP & 1 & 747 & 0.51 & 0.973 & 0.973 & 0.969 & 0.979 & 0.974 & 0.997 \\
C10/MLP & 2 & 747 & 0.51 & 0.975 & 0.974 & 0.962 & 0.990 & 0.976 & 0.997 \\
C10/MLP & 3 & 747 & 0.51 & 0.976 & 0.976 & 0.969 & 0.984 & 0.977 & 0.998 \\
C10/MLP & 5 & 747 & 0.51 & 0.973 & 0.973 & 0.969 & 0.979 & 0.974 & 0.998 \\
FMNIST/ResNet-18 & 1 & 721 & 0.49 & 0.926 & 0.926 & 0.923 & 0.926 & 0.925 & 0.982 \\
FMNIST/ResNet-18 & 2 & 721 & 0.49 & 0.922 & 0.923 & 0.909 & 0.934 & 0.921 & 0.982 \\
FMNIST/ResNet-18 & 3 & 721 & 0.49 & 0.928 & 0.928 & 0.910 & 0.946 & 0.927 & 0.982 \\
FMNIST/ResNet-18 & 5 & 721 & 0.49 & 0.924 & 0.924 & 0.911 & 0.934 & 0.923 & 0.983 \\
FMNIST/SmallCNN & 1 & 733 & 0.52 & 0.940 & 0.940 & 0.945 & 0.940 & 0.942 & 0.989 \\
FMNIST/SmallCNN & 2 & 733 & 0.52 & 0.950 & 0.950 & 0.957 & 0.945 & 0.951 & 0.992 \\
FMNIST/SmallCNN & 3 & 733 & 0.52 & 0.958 & 0.958 & 0.963 & 0.955 & 0.959 & 0.993 \\
FMNIST/SmallCNN & 5 & 733 & 0.52 & 0.945 & 0.945 & 0.948 & 0.948 & 0.948 & 0.992 \\
FMNIST/MLP & 1 & 687 & 0.50 & 0.961 & 0.961 & 0.959 & 0.962 & 0.961 & 0.995 \\
FMNIST/MLP & 2 & 687 & 0.50 & 0.953 & 0.953 & 0.941 & 0.968 & 0.954 & 0.994 \\
FMNIST/MLP & 3 & 687 & 0.50 & 0.962 & 0.962 & 0.957 & 0.968 & 0.962 & 0.993 \\
FMNIST/MLP & 5 & 687 & 0.50 & 0.968 & 0.968 & 0.960 & 0.977 & 0.968 & 0.995 \\
\bottomrule
\end{tabular}
\caption{Relative (high-performance) classification, all four observation
horizons and all six domains, frozen holdout, richest feature set, gradient
boosting. Prevalence is close to but not always exactly 0.50 because the
median threshold is computed on the open 80\% partition while $n$ and
prevalence are measured on the disjoint frozen 20\% partition.}
\label{tab:clffull}
\end{table*}

\begin{table*}[t]
\centering
\scriptsize
\begin{tabular}{lrrrrrrrrr}
\toprule
Domain & $k$ & $n$ & Prev. & Acc. & Bal.Acc. & Prec. & Rec. & F1 & AUC \\
\midrule
C10/ResNet-18 & 1 & 783 & 0.370 & 0.953 & 0.943 & 0.963 & 0.907 & 0.934 & 0.993 \\
C10/ResNet-18 & 2 & 783 & 0.370 & 0.945 & 0.938 & 0.940 & 0.910 & 0.925 & 0.991 \\
C10/ResNet-18 & 3 & 782 & 0.370 & 0.954 & 0.946 & 0.960 & 0.913 & 0.936 & 0.993 \\
C10/ResNet-18 & 5 & 782 & 0.370 & 0.960 & 0.955 & 0.957 & 0.934 & 0.946 & 0.993 \\
C10/SmallCNN & 1 & 756 & 0.345 & 0.940 & 0.936 & 0.906 & 0.923 & 0.915 & 0.986 \\
C10/SmallCNN & 2 & 756 & 0.345 & 0.958 & 0.952 & 0.942 & 0.935 & 0.938 & 0.990 \\
C10/SmallCNN & 3 & 755 & 0.344 & 0.958 & 0.955 & 0.932 & 0.946 & 0.939 & 0.989 \\
C10/SmallCNN & 5 & 754 & 0.344 & 0.956 & 0.952 & 0.935 & 0.938 & 0.936 & 0.991 \\
C10/MLP & 1 & 747 & 0.321 & 0.979 & 0.972 & 0.979 & 0.954 & 0.966 & 0.998 \\
C10/MLP & 2 & 747 & 0.321 & 0.979 & 0.975 & 0.967 & 0.967 & 0.967 & 0.997 \\
C10/MLP & 3 & 747 & 0.321 & 0.985 & 0.983 & 0.979 & 0.975 & 0.977 & 0.998 \\
C10/MLP & 5 & 747 & 0.321 & 0.981 & 0.976 & 0.979 & 0.963 & 0.971 & 0.999 \\
FMNIST/ResNet-18 & 1 & 726 & 0.299 & 0.949 & 0.943 & 0.905 & 0.926 & 0.916 & 0.988 \\
FMNIST/ResNet-18 & 2 & 722 & 0.295 & 0.958 & 0.950 & 0.930 & 0.930 & 0.930 & 0.990 \\
FMNIST/ResNet-18 & 3 & 722 & 0.295 & 0.957 & 0.948 & 0.929 & 0.925 & 0.927 & 0.991 \\
FMNIST/ResNet-18 & 5 & 722 & 0.295 & 0.961 & 0.956 & 0.926 & 0.944 & 0.935 & 0.991 \\
FMNIST/SmallCNN & 1 & 736 & 0.304 & 0.951 & 0.936 & 0.939 & 0.897 & 0.918 & 0.986 \\
FMNIST/SmallCNN & 2 & 736 & 0.304 & 0.967 & 0.956 & 0.963 & 0.929 & 0.945 & 0.989 \\
FMNIST/SmallCNN & 3 & 736 & 0.304 & 0.963 & 0.954 & 0.950 & 0.929 & 0.939 & 0.989 \\
FMNIST/SmallCNN & 5 & 735 & 0.303 & 0.967 & 0.959 & 0.954 & 0.937 & 0.946 & 0.993 \\
FMNIST/MLP & 1 & 689 & 0.219 & 0.962 & 0.933 & 0.943 & 0.881 & 0.911 & 0.993 \\
FMNIST/MLP & 2 & 688 & 0.218 & 0.972 & 0.953 & 0.952 & 0.920 & 0.936 & 0.995 \\
FMNIST/MLP & 3 & 688 & 0.218 & 0.974 & 0.957 & 0.952 & 0.927 & 0.939 & 0.997 \\
FMNIST/MLP & 5 & 687 & 0.217 & 0.975 & 0.960 & 0.952 & 0.933 & 0.942 & 0.997 \\
\bottomrule
\end{tabular}
\caption{Failure prediction, all four observation horizons and all six
domains, frozen holdout, richest feature set, gradient boosting.
Prevalence is the fraction of the eligible cohort labeled a
training-dynamics failure at that horizon; runs already failed before
epoch $k$ are excluded (Section~\ref{sec:protocol}).}
\label{tab:failfull}
\end{table*}

Across all six domains, restricting the failure-prediction cohort to runs
still active at horizon $k$ (rather than all runs with any known terminal
status) changes the eligible sample size by at most 4 runs
(Fashion-MNIST/ResNet-18, $726\to722$ between $k=1$ and $k=5$) and
prevalence by at most 0.004 over the full horizon grid. Most numerical
divergences in this study occur within the very first epoch, before a run
is eligible at any horizon under our leakage-safe rule, so this
eligibility restriction matters more as a safeguard in principle than as a
large active filter in the present data; a study with more later-onset
divergences could see a larger effect from this rule.

\section{Full Transfer Matrix}
\label{app:transfer}

Table~\ref{tab:transferfull} extends the eighteen one-factor-at-a-time
pairs in Figure~\ref{fig:transfer} with the twelve pairs that differ in
both architecture and dataset simultaneously (kind = ``both''), giving the
complete set of thirty ordered off-diagonal domain pairs; the six diagonal
(in-domain) scores are in Table~\ref{tab:headline}. All source models are
gradient boosting, fit on the source domain's open 80\% partition and
evaluated with no retraining on the target domain's frozen 20\% partition;
regression labels are raw final accuracy and classification labels use
each domain's own median (domain-relative), as described in
Section~\ref{sec:transferprotocol}.

\begin{longtable}{llcrrr}
\toprule
Source & Target & Kind & MAE & Spearman & Clf.\ AUC \\
\midrule
\endhead
C10/MLP & C10/ResNet-18 & arch & 18.35 & 0.859 & 0.935 \\
C10/MLP & C10/SmallCNN & arch & 18.54 & 0.819 & 0.961 \\
C10/MLP & FMNIST/MLP & data & 28.52 & 0.828 & 0.869 \\
C10/MLP & FMNIST/ResNet-18 & both & 32.46 & 0.778 & 0.877 \\
C10/MLP & FMNIST/SmallCNN & both & 29.96 & 0.859 & 0.873 \\
C10/ResNet-18 & C10/MLP & arch & 9.44 & 0.604 & 0.716 \\
C10/ResNet-18 & C10/SmallCNN & arch & 6.82 & 0.933 & 0.992 \\
C10/ResNet-18 & FMNIST/MLP & both & 39.07 & 0.638 & 0.904 \\
C10/ResNet-18 & FMNIST/ResNet-18 & data & 31.30 & 0.519 & 0.922 \\
C10/ResNet-18 & FMNIST/SmallCNN & both & 33.59 & 0.436 & 0.938 \\
C10/SmallCNN & C10/MLP & arch & 8.02 & 0.913 & 0.886 \\
C10/SmallCNN & C10/ResNet-18 & arch & 5.11 & 0.969 & 0.988 \\
C10/SmallCNN & FMNIST/MLP & both & 39.90 & 0.812 & 0.929 \\
C10/SmallCNN & FMNIST/ResNet-18 & both & 31.00 & 0.563 & 0.867 \\
C10/SmallCNN & FMNIST/SmallCNN & data & 28.06 & 0.641 & 0.929 \\
FMNIST/MLP & C10/MLP & data & 13.80 & 0.573 & 0.933 \\
FMNIST/MLP & C10/ResNet-18 & both & 11.19 & 0.919 & 0.955 \\
FMNIST/MLP & C10/SmallCNN & both & 10.90 & 0.910 & 0.894 \\
FMNIST/MLP & FMNIST/ResNet-18 & arch & 8.02 & 0.895 & 0.966 \\
FMNIST/MLP & FMNIST/SmallCNN & arch & 6.49 & 0.887 & 0.961 \\
FMNIST/ResNet-18 & C10/MLP & both & 21.38 & 0.802 & 0.810 \\
FMNIST/ResNet-18 & C10/ResNet-18 & data & 10.69 & 0.827 & 0.941 \\
FMNIST/ResNet-18 & C10/SmallCNN & both & 10.33 & 0.813 & 0.841 \\
FMNIST/ResNet-18 & FMNIST/MLP & arch & 9.10 & 0.944 & 0.980 \\
FMNIST/ResNet-18 & FMNIST/SmallCNN & arch & 7.84 & 0.897 & 0.986 \\
FMNIST/SmallCNN & C10/MLP & both & 9.07 & 0.660 & 0.822 \\
FMNIST/SmallCNN & C10/ResNet-18 & both & 12.18 & 0.872 & 0.922 \\
FMNIST/SmallCNN & C10/SmallCNN & data & 10.65 & 0.892 & 0.945 \\
FMNIST/SmallCNN & FMNIST/MLP & arch & 9.05 & 0.950 & 0.974 \\
FMNIST/SmallCNN & FMNIST/ResNet-18 & arch & 4.92 & 0.901 & 0.979 \\
\bottomrule
\caption{Complete off-diagonal transfer matrix, all 30 ordered pairs.
Kind: arch = cross-architecture (dataset fixed), data = cross-dataset
(architecture fixed), both = differs in both factors simultaneously.}
\label{tab:transferfull}
\end{longtable}

\begin{figure}[t]
\centering
\includegraphics[width=\linewidth]{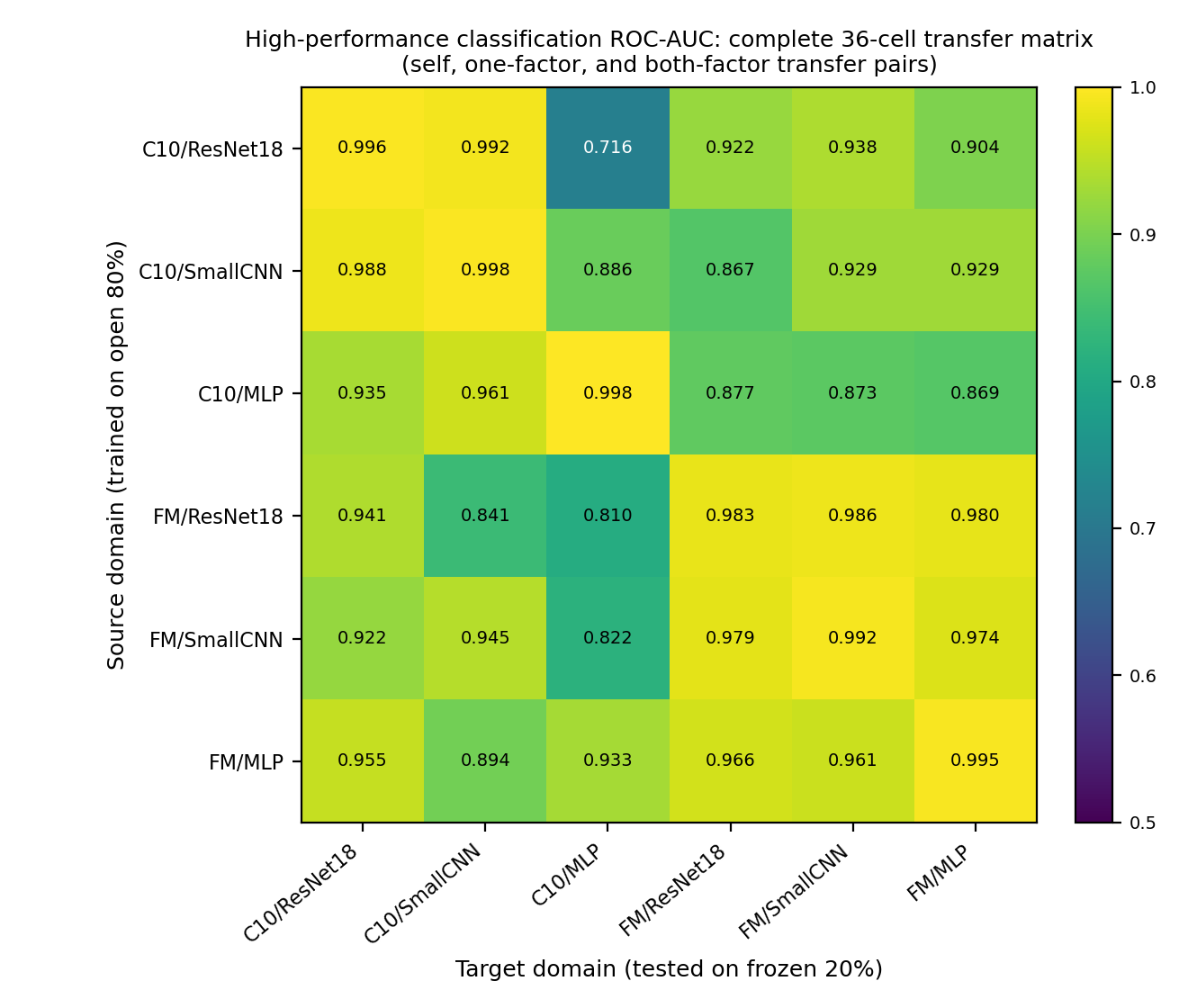}
\caption{Complete 36-cell transfer matrix (18 one-factor pairs, 12
both-differ pairs, and 6 in-domain diagonal scores) as a heatmap; same data
as Table~\ref{tab:transferfull} plus the diagonal.}
\label{fig:fullmatrix}
\end{figure}

\section{Additional Diagnostic Plots}
\label{app:diagnostics}

\begin{figure}[h]
\centering
\includegraphics[width=0.9\linewidth]{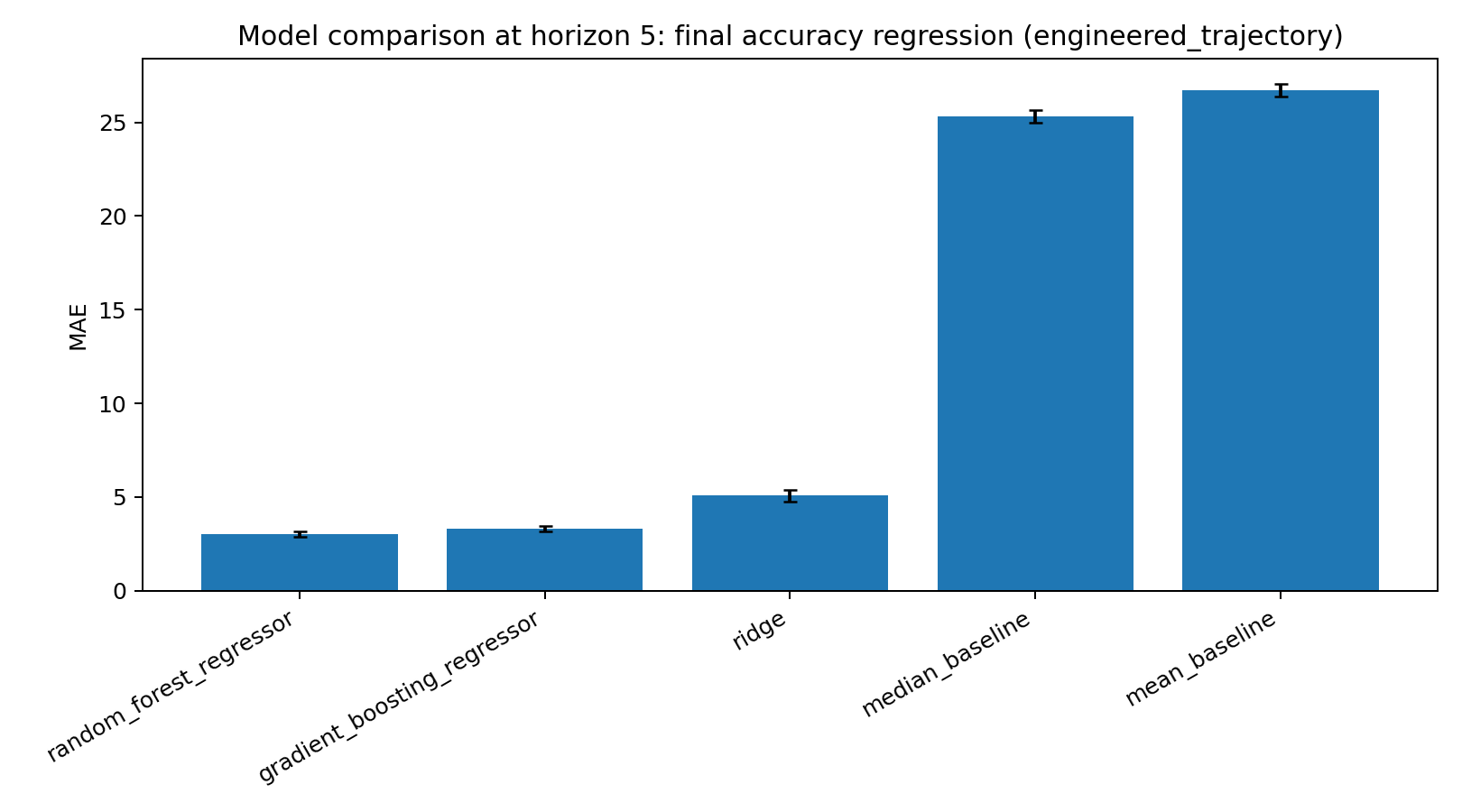}
\caption{Regression baselines and ensembles compared at the five-epoch
horizon (open 80\% development partition, 10 resampled splits),
CIFAR-10/ResNet-18; the feed-forward neural network predictor is evaluated
separately and is not among the models shown here. Gradient boosting was
fixed as the primary model before the frozen evaluation was run; random
forest is retained here as a cross-check rather than a competing final
choice.}
\label{fig:modelcomp}
\end{figure}

\begin{figure}[h]
\centering
\includegraphics[width=0.7\linewidth]{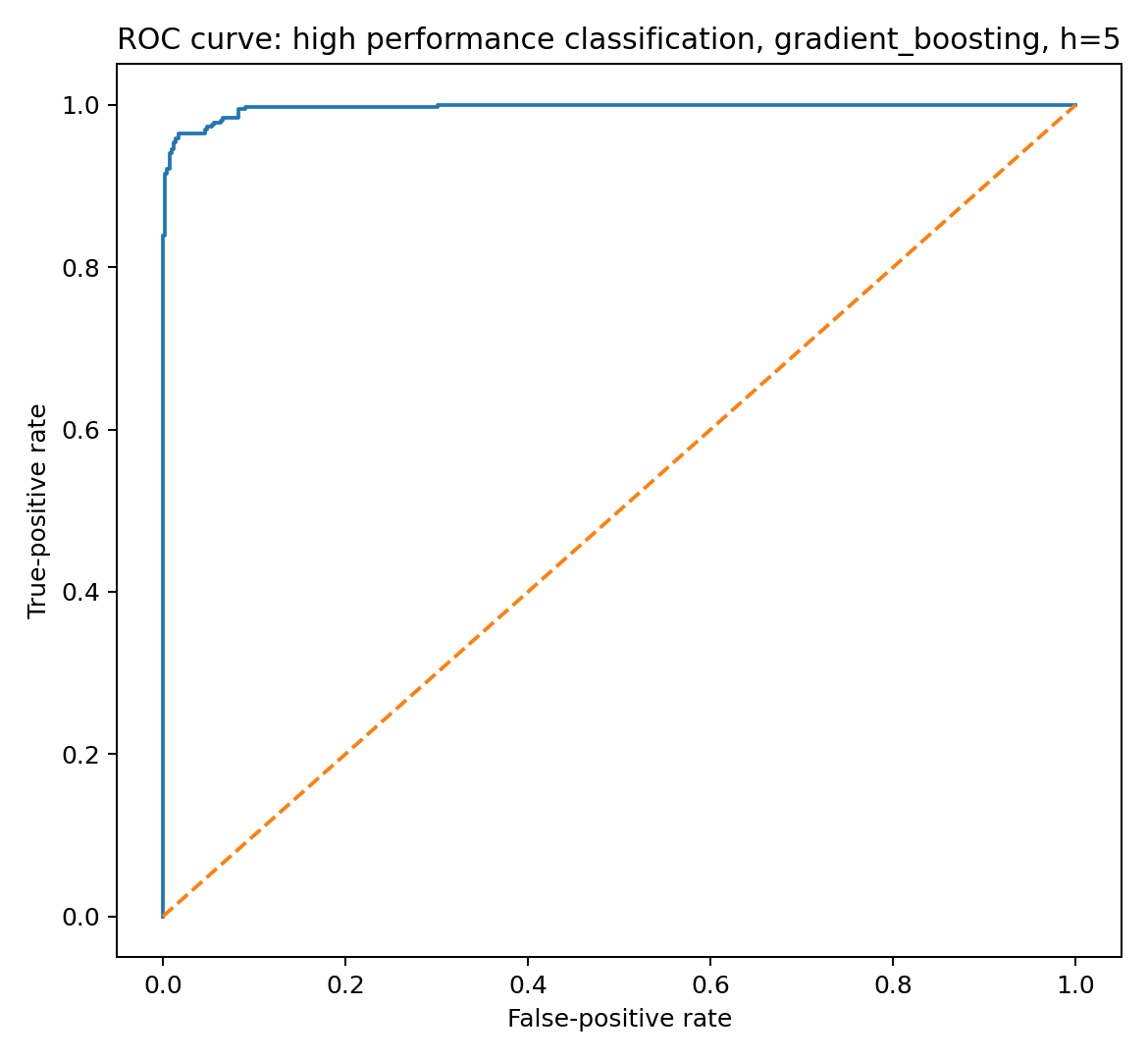}
\caption{ROC curve, relative classification, five-epoch horizon,
CIFAR-10/ResNet-18, frozen holdout.}
\label{fig:clfroc}
\end{figure}

\begin{figure}[h]
\centering
\includegraphics[width=0.7\linewidth]{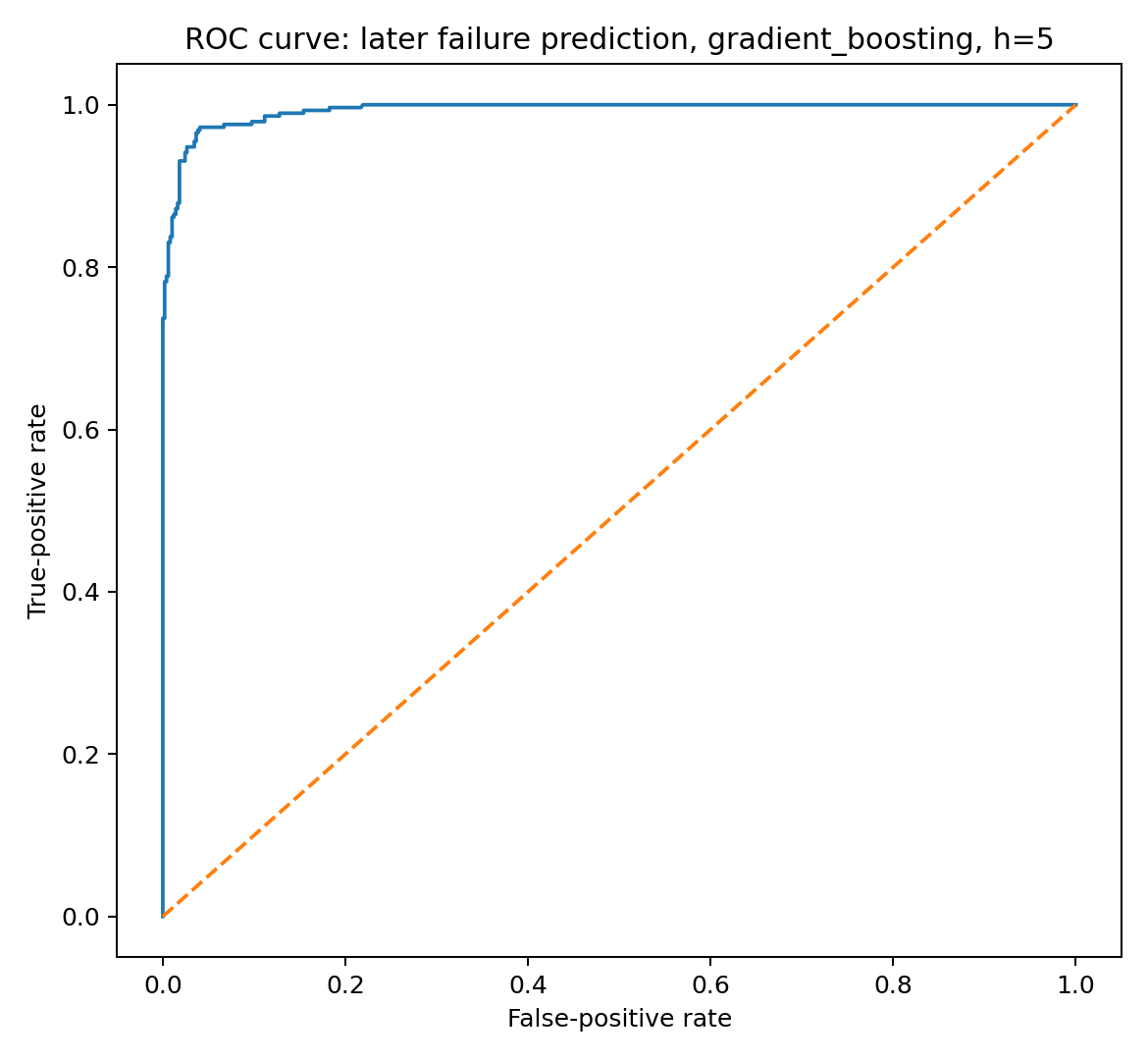}
\caption{ROC curve, failure prediction, five-epoch horizon,
CIFAR-10/ResNet-18, frozen holdout.}
\label{fig:failroc}
\end{figure}

Figure~\ref{fig:modelcomp} compares model families beyond gradient boosting
and the feed-forward neural network discussed in
Section~\ref{sec:indomain}; Figures~\ref{fig:clfroc}
and~\ref{fig:failroc} give the ROC curves underlying the relative-classification
and failure-prediction ROC-AUC values reported for CIFAR-10/ResNet-18 in
Section~\ref{sec:indomain}.

\section{Confidence-Interval Methodology}
\label{app:ci}

For the repeated grouped-holdout evaluation used during model development
(Section~\ref{sec:evalsplits}), we draw 10 grouped train/test splits at a
25\% test fraction, grouped by hyperparameter configuration, for each
(endpoint, horizon, feature set, model) combination. Writing $\bar{x}$ for
the mean of the resulting 10 per-split metric values and SE for their
ordinary standard error, we report a 95\% interval as
\[
\bar{x} \;\pm\; t_{0.975,\,n-1}\cdot\text{SE}\cdot\sqrt{1+n_{\text{test}}/n_{\text{train}}},
\]
where $n=10$ is the number of splits. The factor
$\sqrt{1+n_{\text{test}}/n_{\text{train}}}$ is the Nadeau--Bengio
correction, which accounts for the fact that repeated splits drawn from a
fixed, overlapping pool of training data are positively correlated with
each other; without this correction, the naive standard error across splits
becomes anti-conservative (artificially narrow simply from increasing the
number of resampled splits, rather than from any real increase in
information). All intervals are clipped to each metric's admissible range
(for example, $[0,1]$ for ROC-AUC). This interval characterizes
split-to-split variability within our one fixed set of sampled
hyperparameter configurations; it is not a generalization bound to an
independently drawn new population of configurations, which is instead
what the frozen-holdout evaluation (Section~\ref{sec:frozenholdout})
addresses, at the cost of providing only a single point estimate rather
than an interval. A direct consequence: small non-monotonic reversals
between adjacent horizons in the frozen-holdout tables
(Appendix~\ref{app:metrics}) -- for example, Fashion-MNIST/MLP regression
$R^2$ dips from $0.970$ at $k=2$ to $0.965$ at $k=3$ before rising to
$0.982$ at $k=5$ -- are ordinary point-estimate noise on a single frozen
split, not evidence that additional telemetry ever degrades prediction.

\section{Data Accounting and Quality Checks}
\label{app:dataaccounting}

Table~\ref{tab:status} gives the complete terminal-status breakdown behind
Table~\ref{tab:accounting}. ``Success'' indicates final accuracy at or
above the domain's usability threshold; ``Failed'' indicates the run
completed all 15 epochs below that threshold; ``NaN failure'' indicates
numerical divergence. No run in any domain's final data terminated in an
uncaught infrastructure error; we confirmed this by inspecting every raw
data export directly rather than inferring it from the deduplicated output
alone.

\begin{table}[H]
\centering
\small
\begin{tabular}{lrrr}
\toprule
Domain & Success & Failed & NaN failure \\
\midrule
C10 / ResNet-18    & 2{,}519 (63.0\%) & 1{,}355 (33.9\%) & 124 (3.1\%) \\
C10 / SmallCNN     & 2{,}403 (60.1\%) & 1{,}340 (33.5\%) & 257 (6.4\%) \\
C10 / MLP          & 2{,}443 (61.1\%) & 1{,}207 (30.2\%) & 350 (8.8\%) \\
FMNIST / ResNet-18 & 2{,}520 (66.5\%) & 1{,}084 (28.6\%) & 186 (4.9\%) \\
FMNIST / SmallCNN  & 2{,}505 (62.6\%) & 1{,}123 (28.1\%) & 372 (9.3\%) \\
FMNIST / MLP       & 2{,}632 (65.8\%) & 764 (19.1\%) & 604 (15.1\%) \\
\bottomrule
\end{tabular}
\caption{Terminal-status breakdown, final deduplicated data. Row totals
match the ``Final'' column of Table~\ref{tab:accounting}.}
\label{tab:status}
\end{table}

The usability thresholds (Table~\ref{tab:accounting}) were fixed before
generating any data: ResNet-18/CIFAR-10 45\%, SmallCNN/CIFAR-10 40\%,
MLP/CIFAR-10 30\% (roughly half of a well-tuned run's accuracy on the
harder dataset), and ResNet-18/Fashion-MNIST 80\%, SmallCNN/Fashion-MNIST
78\%, MLP/Fashion-MNIST 70\% (closer to that dataset's achievable ceiling,
since Fashion-MNIST is easy enough that a literal half-of-ceiling threshold
would rarely be crossed at all). Because the thresholds sit at different
relative positions on the two datasets' achievable accuracy ranges,
``failure'' is effectively a stricter bar on Fashion-MNIST than on
CIFAR-10; this is worth keeping in mind when comparing failure-prediction
ROC-AUC (Table~\ref{tab:headline}, Table~\ref{tab:failfull}) across
datasets. No threshold was ever estimated from evaluation data at any
point in the pipeline.

\textbf{Duplicate-identity forensics, in full.} The naive per-row
identifier assigned during data collection restarts numbering within each
sampling phase and is therefore reused across phases; deduplicating on it
directly would incorrectly merge unrelated runs from the two phases that
happen to share a row number. We instead deduplicate on a hash of the
sampled hyperparameter tuple, which is stable across phases, and separately
verified that no domain contains two rows that share this hash but differ
in their recorded random training seed (which would indicate a true
identity collision rather than an artifact of the naive identifier). The
Fashion-MNIST/ResNet-18 domain additionally had two data exports covering
an identical set of 200 run identifiers at different levels of file
granularity; of the 200 corresponding row pairs, 40 were byte-identical and
160 shared identifiers but disagreed in their recorded telemetry values, a
pattern consistent with ordinary GPU/cuDNN non-determinism across two
separate executions of the same configuration, and concentrated
disproportionately near the second sampling phase's success/failure
boundary. Rather than resolve the 160 disagreeing rows with an arbitrary
row-level tie-break, we removed the entire redundant finer-granularity
export, which is the 200-row removal in Table~\ref{tab:accounting}.

\textbf{Raw row-count shortfall.} Two domains (CIFAR-10/ResNet-18 and
Fashion-MNIST/ResNet-18) fall 2 and 10 runs short, respectively, of the
4{,}000-run target reached in every other domain, even before the
200-row duplicate removal described above. We verified that every raw row
in both domains carries a genuine terminal status (Success, Failed, or NaN
failure) with zero rows of any uncaught-error status among them; the
shortfall therefore reflects a small number of sampled configurations for
which the training-run scheduler produced no output row at all -- for
example, a job killed by the cluster scheduler before the training process
could begin -- rather than a completed run whose failure status was
silently dropped from the final data.

\end{document}